\documentclass{article}

\usepackage{microtype}
\usepackage{graphicx}
\usepackage{wrapfig}
\usepackage{caption}
\usepackage{subcaption}
\usepackage{url}
\usepackage{tabu}
\usepackage{longtable}
\usepackage{xspace}
\usepackage{enumitem}
\usepackage{titletoc}
\usepackage{listings}
\usepackage{color}
\usepackage{booktabs} 

\definecolor{codegreen}{rgb}{0,0.6,0}
\definecolor{codegray}{rgb}{0.5,0.5,0.5}
\definecolor{codepurple}{rgb}{0.58,0,0.82}
\definecolor{backcolour}{rgb}{0.95,0.95,0.92}

\lstdefinestyle{mystyle}{
    backgroundcolor=\color{backcolour},   
    commentstyle=\color{codegreen},
    keywordstyle=\color{magenta},
    numberstyle=\tiny\color{codegray},
    stringstyle=\color{codepurple},
    basicstyle=\ttfamily\footnotesize,
    breakatwhitespace=false,         
    breaklines=true,                 
    captionpos=b,                    
    keepspaces=true,                 
    numbers=left,                    
    numbersep=5pt,                  
    showspaces=false,                
    showstringspaces=false,
    showtabs=false,                  
    tabsize=4
}
\usepackage{hyperref}

\usepackage[accepted]{icml2024}

\usepackage{amsmath}
\usepackage{amssymb}
\usepackage{mathtools}
\usepackage{amsthm}

\usepackage[capitalize,noabbrev]{cleveref}

\theoremstyle{plain}

\theoremstyle{definition}

\theoremstyle{remark}

\usepackage[textsize=tiny]{todonotes}
\newcommand{\method}{\texttt{UniPredict}\xspace}

\icmltitlerunning{Submission and Formatting Instructions for ICML 2024}

\begin{document}

\twocolumn[
\icmltitle{\method: Large Language Models are Universal Tabular Classifiers}



\icmlsetsymbol{equal}{*}

\begin{icmlauthorlist}
\icmlauthor{Ruiyu Wang}{equal,UofT}
\icmlauthor{Zifeng Wang}{equal,UIUC}
\icmlauthor{Jimeng Sun}{UIUC}

\end{icmlauthorlist}

\icmlaffiliation{UofT}{Department of Computer Science, University of Toronto}
\icmlaffiliation{UIUC}{University of Illinois Urbana-Champaign}

\icmlcorrespondingauthor{Ruiyu Wang}{rwang@cs.toronto.edu}
\icmlcorrespondingauthor{Zifeng Wang}{zifengw2@@illinois.edu}

\icmlkeywords{Machine Learning, ICML, Large Language Models, Tabular Learning}

\vskip 0.3in
]



\printAffiliationsAndNotice{\icmlEqualContribution} 

\begin{abstract}
Tabular data prediction is a fundamental machine learning task for many applications. Existing methods predominantly employ discriminative modeling and operate under the assumption of a fixed target column, necessitating re-training for every new predictive task.  Inspired by the generative power of large language models (LLMs), this paper exploits the idea of building universal tabular data predictors based on generative modeling, namely \method. Here, we demonstrate the scalability of an LLM to extensive tabular datasets, enabling it to comprehend diverse tabular inputs and predict target variables following the provided instructions. Specifically, we train a single LLM on an aggregation of 169 tabular datasets with diverse targets and compare its performance against baselines that are trained on each dataset separately. We observe this versatile \method model demonstrates an advantage over other models, ranging from 5.4\% to 13.4\%, when compared with the best tree-boosting baseline and the best neural network baseline, respectively. We further test \method in few-shot learning settings on another 62 tabular datasets. Our method achieves strong performance in quickly adapting to new tasks. In low-resource few-shot setup, we observed a 100\%+ performance advantage compared with XGBoost, and significant margin over all baselines. We envision that \method sheds light on developing a universal tabular data prediction system that learns from data at scale and serves a wide range of prediction tasks.
\end{abstract}

\section{Introduction}
Tabular data is organized in a tabular or spreadsheet format within a relational database. Each row within the table corresponds to a specific data sample, and the columns encompass a range of feature variables with diverse types, such as categorical, numerical, binary, and textual features. Tabular data prediction is fundamental to many real-world machine-learning applications such as click-through rate prediction~\citep{yang2022click} and medical outcome prediction~\citep{wang2022transtab}.

Nonetheless, most previous methods fall short by assuming a \textit{fixed target}. This entails selecting a specific column, such as patient mortality in breast cancer cases, with the other columns as the input features. Therefore, a model trained to predict this particular target becomes specialized and cannot be employed for predicting any other target, such as cancer relapse. To predict a different target, one must create a new dataset corresponding to the desired target and retrain the model. This practice renders substantial work involved in developing and hosting dataset-specific tabular data predictors. 

\begin{figure*}[htbp]
     \centering
     \begin{subfigure}[b]{0.305\linewidth}
         \centering
         \includegraphics[width=\linewidth]{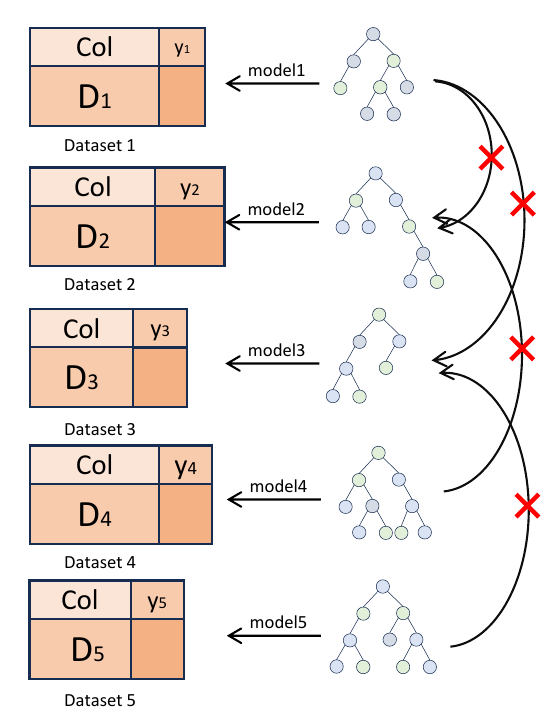}
         \caption{Traditional tabular models}
         \label{fig:1a}
     \end{subfigure}
     \begin{subfigure}[b]{0.305\linewidth}
         \centering
         \includegraphics[width=\linewidth]{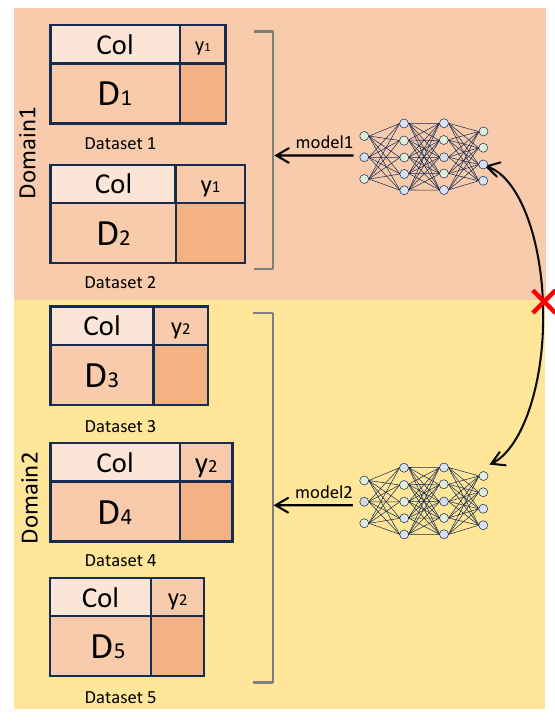}
         \caption{In-domain tabular models}
         \label{fig:1b}
     \end{subfigure}
     \begin{subfigure}[b]{0.36\linewidth}
         \centering
         \includegraphics[width=\linewidth]{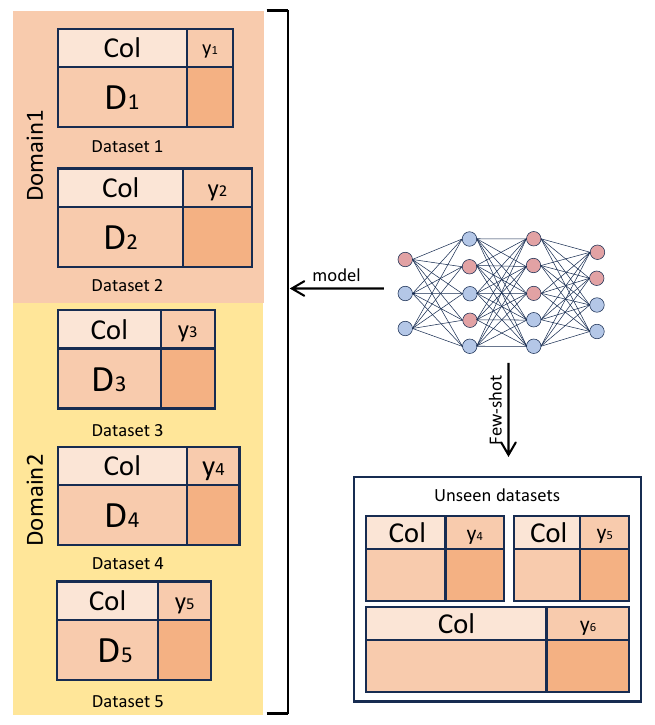}
         \caption{Universal tabular models}
         \label{fig:1c}
     \end{subfigure}
        \caption{Visualization for three tabular modeling paradigms. \textbf{Left}: In \texttt{Traditional} Tabular Modeling tasks (Figure \ref{fig:1a}), distinct models are trained individually on each dataset, making them incapable of adaptation to new datasets with differing features and targets. \textbf{Middle}: In the \texttt{In-Domain} Tabular Modeling tasks (Figure \ref{fig:1b}), where flexibility is allowed for features, the targets remain the same across datasets. \textbf{Right}: the proposed \texttt{Universal} Tabular Modeling paradigm (Figure \ref{fig:1c}), which accommodates arbitrary inputs and predicting for arbitrary targets. This paradigm does not impose any restrictions on the domains of the datasets used. In \texttt{Universal} Tabular Modeling, the datasets can originate from entirely different domains.}
        \label{fig:comparisons}
\end{figure*}

Unlike most traditional algorithms that make \textit{discriminative} modeling for tabular prediction, we intend to harness LLMs for tabular prediction through \textit{generative} modeling. Figure \ref{fig:comparisons} demonstrates the difference between the previous practices and our modeling paradigm. This paradigm provides substantial flexibility in (1) processing natural language descriptions of tabular data and (2) generating predictions for specified target labels based on input instructions. While previous works have tried to fine-tune LLMs for generating target labels of tabular data \citep{dinh2022lift,hegselmann2023tabllm}, they have their limitations, mainly in that they still require training specific predictors for each dataset and target variable. Moreover, these generative prediction methods do not provide the associated confidence of their predictions as traditional tabular prediction models do. By contrast, the goal of this work is to build universal tabular predictors based on generative LLM, which accept \textit{arbitrary} inputs and predict for \textit{arbitrary} targets, following the input instructions.

Specifically, this work explores the ways to unlock the potential of LLMs as universal tabular data predictors, namely \method, which hinges on the following insights: 

\begin{itemize}[leftmargin=*]
    \item \textbf{Data Scale}: Scaling to 160+ diverse tabular datasets to fuel the training of a powerful LLM that performs prediction for diverse inputs and targets.
    \item \textbf{Prompt Engineering}: The prompts that integrate the metadata (e.g., the dataset description and schema of columns), the input tabular sample, and the instruction for prediction generation.
    \item \textbf{Instruction Tuning}: Instruction tuning that encourages LLM to not only generate the label but also provide confidence estimates for its predictions.
\end{itemize} 


We elaborate on our framework in Section \ref{sec:method}, which is followed by the experiment results in Section~\ref{sec:experiment}. In detail, we train a single \method model on the aggregated training sets from 169 tabular datasets and test it on the corresponding test sets. For comparison, we train one unique baseline model for each tabular dataset and report their performances. We observe that the universal tabular predictor \method outperforms the best neural network baselines by $13.4\%$ and the best boosting algorithms by $5.4\%$, across the test sets. Additionally, we observed that \texttt{UniPredict} exhibits an advantage in the low-resource regime. Even as the sample size increases, it consistently maintains among the top models. We close with the discussion of related papers in Section \ref{sec:related-works} and the conclusion in Section \ref{sec:conclusion}.


\section{Method and Implementation}\label{sec:method}

\begin{figure*}[h]
\begin{center}
\includegraphics[width=0.8\textwidth]{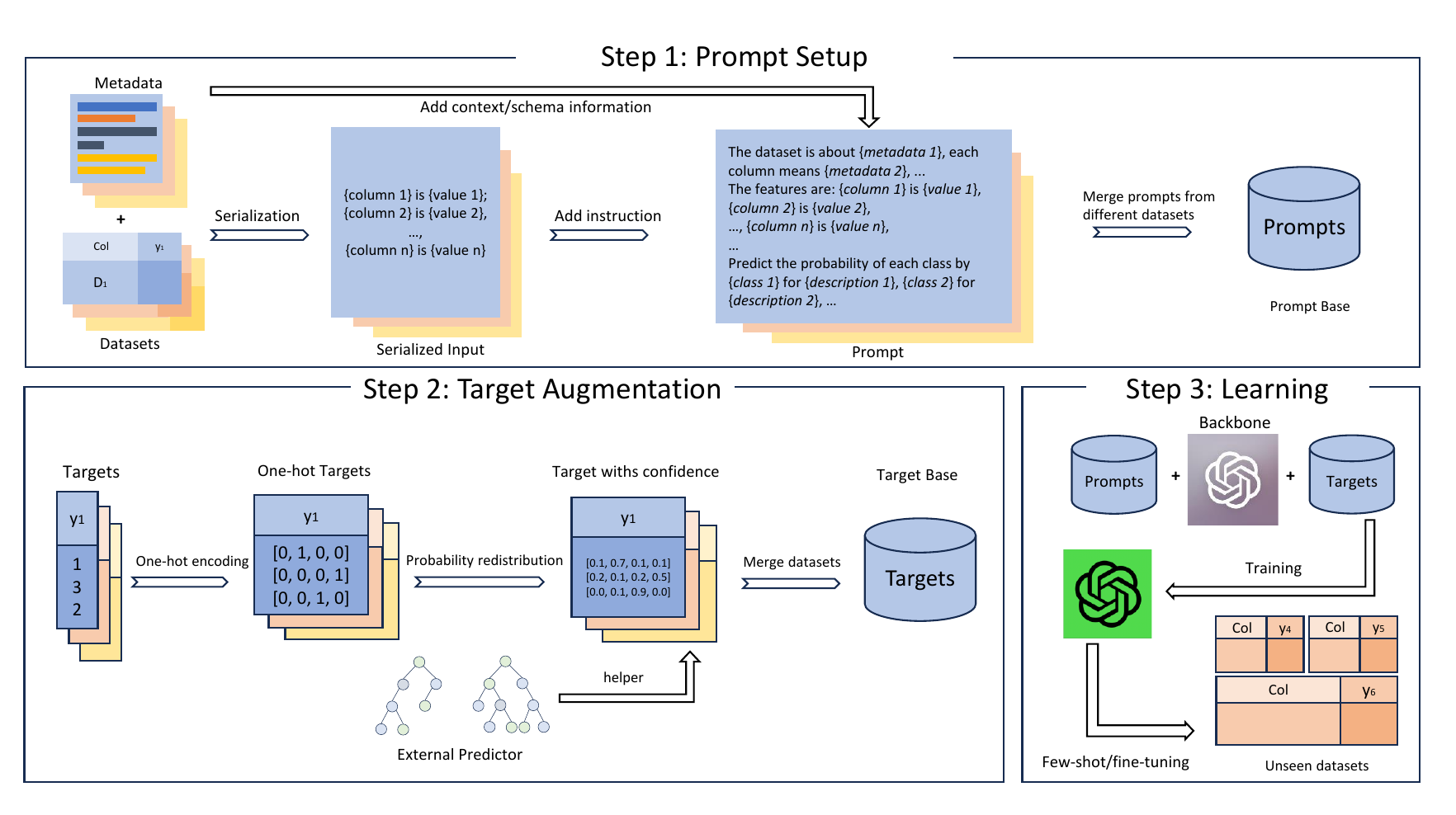}
\end{center}
\caption{The \method framework. It consists of three steps: 1) \textbf{Prompt Setup} sets up the prompts by metadata, sample serialization, and instructions; 2) \textbf{Target Augmentation} transforms target values into categories with confidence estimates; and 3) \textbf{Learning} fine-tunes the backbone model by prompts and targets yielded from the previous procedures.}
\label{fig:framework-demo}
\end{figure*}

\subsection{Problem Formulation}
Before going into details of the proposed method, we define two problems that we aim to resolve:

\paragraph{Universal Tabular Modeling} Given a dataset $\mathbf{D}_n$ in \textit{any} domain, we have its components $\mathbf{D}_n = \{\mathbf{M}_n, \mathbf{S}_n;\mathbf{T}_n\}$ that include the metadata $\mathbf{M}_n$, samples $\mathbf{S}_n$, and targets $\mathbf{T}_n$.
Different from traditional tabular models  $\displaystyle f_n: \mathbf{S}_n \rightarrow \mathbf{T}_n$ (shown in Figure \ref{fig:1a}) that gives a 1-to-1 dataset-model relationship, or in-domain tabular models $\displaystyle f_{task}: \mathbf{S}_n \rightarrow \mathbf{T}_{task}$ (shown in Figure \ref{fig:1b}), we require a universal model $\displaystyle f_{univ}: \mathbf{S} \rightarrow \mathbf{T}$ such that $\displaystyle f_{univ}(\mathbf{S}_n; \mathbf{M}_n) = \mathbf{T}_n$.
This approach enables us to create a more versatile prediction setting. The model parameters are no longer dependent on any particular dataset or task domain. Instead, a single set of parameters, with the aid of metadata, can be used for all datasets from any domain (shown in Figure \ref{fig:1c}).

\paragraph{Few-shot Learning} We expect our model $\displaystyle f$ that is trained on datasets $\{\mathbf{D}_1, \mathbf{D}_2, \cdots \mathbf{D}_n \}$ to be also available to predict for a new target $\mathbf{T}_{n+1}$, given $(\mathbf{S}_{n+1}, \mathbf{M}_{n+1}) \in \mathbf{D}_{n+1}$. 
We can fine-tune $f$ with the new dataset $\mathbf{D}_{n+1}$ in a low-resource regime to achieve few-shot learning.

As illustrated in Figure \ref{fig:framework-demo}, The \method framework is structured around three primary steps: First, in \textbf{Prompt Setup} \S\ref{sec:prompt}, prompts are established through metadata, sample serialization, and instructions. Second, \textbf{Target Augmentation} \S\ref{sec:augmentation} involves transforming target values into categorized forms accompanied by confidence estimates. Last, the \textbf{Learning} \S\ref{sec:learning} step fine-tunes the backbone model utilizing the prompts and targets derived from the preceding procedures.

\subsection{Prompt Engineering} \label{sec:prompt}
Tabular data have to be transformed into natural language inputs to be comprehended by LLMs. It is highlighted that the quality of this natural language input has a major impact on the LLM's performance \citep{zhao2021calibrate}. We hereby present how we formulate the input prompt for our \method framework. Technically, based on dataset $\mathbf{D} = \{\mathbf{M}, \mathbf{S};\mathbf{T}\}$ we define the function \texttt{prompt}($\hat{\mathbf{M}}, \hat{\mathbf{S}}, \mathbf{I}$) that takes pre-processed metadata $\hat{\mathbf{M}}$ and tabular sample $\hat{\mathbf{S}}$, and the instruction $\mathbf{I}$ as input and perform serialization to produce the natural language input for LLMs:

\textbf{Metadata} $\hat{\mathbf{M}}$ represents a \textit{serialized} description of the context and schema definition of the dataset.

\textbf{Tabular Sample} $\hat{\mathbf{S}}$ that represents \textit{serialized} contents of the raw sample.

\textbf{Instruction} $\mathbf{I}$ that contains the guidance that prompts LLMs to make the final prediction about the target, e.g., the probability prediction for each target class.


We describe the detailed setup of these components in the following sections. We also offer the example of used prompts in Appendix~\ref{apdx:prompt}.

\paragraph{Metadata Re-formatting}  \label{sec:metadata}
As \method accommodates a wide range of tabular datasets that share distinct schema, the dataset metadata plays a vital role in facilitating the language modeling on these diverse tabular data. For instance, many table columns are abbreviations or coded with a private dictionary, thus hurdling LLMs in comprehending the tabular inputs. In practice, the metadata is usually provided in unstructured texts with the raw dataset. Here, we propose to design a function \texttt{reformat($\mathbf{M}$)} that consolidates arbitrary input $\mathbf{M}$ to (1) a description of the target to predict and (2) the semantic descriptions of features. We employ GPT-3.5\footnote{OpenAI API: \texttt{gpt-3.5-turbo}} to automate the metadata reformatting process. We offer the example metadata reformatting process in Appendix \ref{apdx:metadata}.

\paragraph{Feature Serialization} \label{sec:feature-serialization}

Given the raw metadata $\mathbf{M}$ and the samples $\mathbf{S}$, we define the function \texttt{serialize}$(c, v)$ to produce a \texttt{str} output given the column names \texttt{c} and feature values \texttt{v}, where \texttt{c} $\in$ \texttt{reformat($\mathbf{M}$)} and \texttt{v} $\in \mathbf{S}$.
Each value is paired with the corresponding column in the format of ``\textit{\{column\} is \{value\}, \{column\} is \{value\}, ...}". Besides, we round numeric values to a fixed precision before tokenization, and more data-dependent binning methods, such as adaptive histogram binning, may be considered. Some examples of the serialization can be found in Appendix \ref{apdx:feature}. 

\subsection{Instruction Formulation \& Target Augmentation}\label{sec:augmentation}
When encountering tabular data prediction with LLM, the most natural idea is to put the tabular sample as the input and prompt LLM to generate the target label \citep{dinh2022lift,hegselmann2023tabllm}. For instance, prompting LLM with the input ``\textit{Is the person's annual income $\geq 50$?}" to yield the output ``\textit{yes}" or ``\textit{no}" as the binary prediction. However, it has two main drawbacks:
\begin{itemize}[leftmargin=*]
    \item \textbf{Reliability} Unlike traditional ML algorithms that produce the probability prediction for each class, this method merely produces the output label. Due to the uncertainty in text generation, the label prediction from LLM may be unreliable without a numerical estimation of its confidence.
    \item \textbf{Robustness} We empirically identified this modeling paradigm may fail to converge when encountering challenging tabular prediction tasks or noisy inputs. In these scenarios, the LLM may either refuse to generate predictions or tend to continue the input texts.
\end{itemize}

To overcome these challenges, we propose instructing models to predict each target class probability, e.g., ``\textit{yes: 0.8; no: 0.2}". This is achieved by adding another \textit{target augmentation} step. 


\paragraph{Target Augmentation} \label{sec:target-augmentation}
We transform the target label into a set of probabilities for each class via a function called ``augment". Formally, for target $\mathbf{T}$ in an arbitrary dataset $\mathbf{D}$, we define a function $\texttt{augment}(\mathbf{T})=\{\mathbf{C},\mathbf{P}\}$, where $\mathbf{C}$ are new categories of targets with semantic meaning and $\mathbf{P}$ are the assigned probabilities to each category. We extend the target into categorical one-hot encoding and then use an \textit{external predictor} to create the calibrated probability distributions. This replaces the 0/1 one-hot encoding while maintaining the final prediction outcome. For datasets with discrete target values (e.g., classification), the target classes are processed by one-hot encoding. For continuous numerical targets (e.g., regression), the categories are defined by their quantiles.

We use an isotopic calibrated XGBoost classifier \citep{chen2016xgboost} with \texttt{ n\_estimators=100 } as the external predictor. We train one predictor for each dataset and then leverage it to produce the probability for each class for all samples.
It is noted that this predictor serves as a probability estimator for sample labels without the loss of information or data leakage. Formally, given the target classes \texttt{t} $\in \{ 0, ..., |\mathbf{C}|\}$ and target probabilities \texttt{p} $\in \mathbf{P}$, we define a function \texttt{serialize\_target(t, p)} that serializes target classes and probabilities into a sequence formatted as ``\texttt{class} $\{t_1\}:\{p_1\}$, \texttt{class} $\{t_2\}:\{p_2\}$, $\dots$". 
This sequence is used as the referenced output to fine-tune the LLM. Besides the merit of entailing confidence predictions, target augmentation offers more sufficient supervision for LLMs, which we find vital for its robustness in training and inference.


\paragraph{Instruction Formulation}
The instruction $\mathbf{I}$ describes the objective that prompts LLM to comprehend the input tabular sample and predict for the augmented target $\texttt{augment}(\mathbf{T})$. Given the target classes \texttt{t} $\in [ 0, |\mathbf{C}|]$ and target semantic explanation \texttt{e} $\in \mathbf{C}$, we define a function \texttt{serialize\_class(t, e)} that converts the classes \texttt{t}, and their corresponding semantic explanation \texttt{e}, into a natural language sequence ``$\texttt{class} \ \{t\} \ \texttt{means} \ \{e\}, \dots$". We present the example prompts in Appendix \ref{apdx:target}.



\subsection{Learning}\label{sec:learning}
\paragraph{LLM for Tabular Prediction}
During fine-tuning, our objective is to minimize the difference between the output sequence generated by the adapted LLM function (represented by LLM(prompt($\hat{\mathbf{M}}, \hat{\mathbf{S}}, \mathbf{I}$))) and the reference output sequence generated from target augmentation (represented by serialize\_target(augment($\mathbf{T}$))). 
However, during testing, we evaluate the prediction correctness instead of the similarity between the output and reference sequences. To do this, we map the natural language sequence generated by the LLM function to the actual class that the model is predicting. We then check the correctness of the prediction by comparing it with the ground truth label. 
We use a regex expression matching technique for the mapping procedure. We have included examples for such comparisons in Appendix \ref{apdx:output-mapping}.

\paragraph{Learning} In our model learning process, we generate prompts using samples and metadata from different datasets and update the model based on instruction fine-tuning. 
Subsequently, we assess the model's actual performance by comparing its class predictions (after output mapping) to the original target values. This evaluation is conducted on both the datasets used during training and previously unseen datasets.
We adapt GPT-2 \citep{radford2019language} as our backbone, and we used the huggingface\footnote{https://huggingface.co/} package for training. See Appendix \ref{apdx:training} for the detail of parameter choice.

\subsection{Our Implementation of UniPredict} \label{sec:implementation}
\paragraph{Dataset Setup}
We collect the datasets from Kaggle\footnote{https://www.kaggle.com/datasets/}.
We pre-select the datasets from the \texttt{classification} category and drop the datasets that do not provide organized and recognizable metadata.
We leverage the \texttt{Kaggle API} \footnote{https://github.com/Kaggle/kaggle-api} to download both the raw data and their descriptions with an argument \texttt{--file-size csv} to restrict the dataset format.
In this way, we simplify the follow-up dataset reading procedures.
To ensure a comprehensive evaluation, we do not preselect datasets by their domains, categories, or purposes.

We end with the training corpus built from 169 datasets. For each selected dataset, we perform a max-size cutoff at 7500 samples to prevent any datasets with too many samples from dominating the corpus. The number of training samples in the entire corpus is 366,786. Dataset statistics can be found in Appendix \ref{apdx:dataset-statistics}.

\paragraph{Implementations}
The target augmentation step is done by the \texttt{XGBoost}  classifiers. However, as mentioned in Section \ref{sec:target-augmentation}, we accept other classifiers to be adapted as long as they produce proper probability values.
Furthermore, measuring the information entailed by different classifiers in this problem is also a potential topic to explore.

We utilize a GPT-2 \citep{radford2019language} model as backbone. Besides the normal \texttt{UniPredict} framework, we instantiate a variant that only takes feature names from the metadata, named as \texttt{UniPredict-light}; in contrast, we named our normal version \texttt{UniPredict-heavy}. \texttt{UniPredict-light} is expected to take less time for fine-tuning and demonstrate an equal or better performance when the dataset is well-maintained. Since no assumptions should be made to unknown datasets, \texttt{UniPredict-heavy} is the most reliable baseline.
The difference in implementation of the two variants can be found in Appendix \ref{apdx:prompt}.

\section{Experiment}\label{sec:experiment}
In this section, we conducted extensive experiments with \method and a suite of cutting-edge tabular prediction baselines, with a focus on answering the following research questions:

\begin{itemize}[leftmargin=*]
    \item \textbf{Universal Tabular Modeling} (Section \ref{sec:training-results}) Can a single \method model succeed in performing a universal modeling of extensive tabular datasets?
    \item \textbf{Few-shot learning} (Section \ref{sec:few-shot-results}) Compared with the baselines, how well does a pre-trained \method model adapt to new tasks?
    \item \textbf{Analysis \#1} (Section \ref{sec:case-study}) Under what circumstances is \method less competitive to others?
    \item \textbf{Analysis \#2} (Section \ref{sec:ablation}) What are the key factors that make \method a successful candidate for universal tabular prediction?
\end{itemize}

\subsection{Baseline Models} \label{sec:baseline}
We included \texttt{MLP} as the simplest neural baseline. Drawing inspiration from the effectiveness of tree-boosting algorithms on tabular tasks, we assessed the performance of \texttt{XGBoost} \citep{chen2016xgboost}, a preeminent model in this domain.
To explore the effectiveness of attention-based models in our tasks, we also included \texttt{TabNet} \citep{arik2021tabnet} and \texttt{FT-Transformer} \citep{gorishniy2021revisiting} to our experimental evaluation. Additionally, we incorporated \texttt{TabLLM} \citep{hegselmann2023tabllm} into our analysis, as it represents another model designed for tabular data with a focus on Large Language Models. The configurations and specifics of these baseline models are provided in Appendix \ref{apdx:baseline}.
Given the dataset-specific and non-transferable nature of the baseline models, we established isolated instances for each dataset included in our study. In contrast, for \texttt{UniPredict}, which aims at Universal Tabular Prediction, we instantiated a single model instance capable of handling all the datasets used in our experimentation.

\subsection{Results on Universal Tabular Modeling}
\label{sec:training-results}

\begin{figure*}
     \centering
     \begin{subfigure}[b]{0.45\textwidth}
         \centering
         \includegraphics[width=\textwidth]{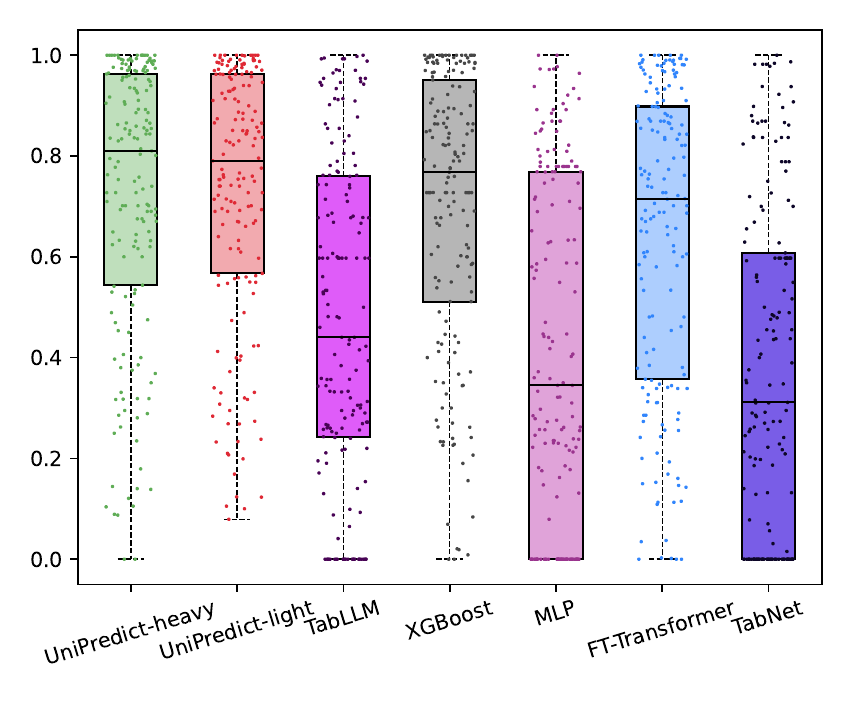}
         \caption{Model Accuracy}
         \label{fig:supervised-dataset-result-accuracy}
     \end{subfigure}
     \hfill
     \begin{subfigure}[b]{0.45\textwidth}
         \centering
         \includegraphics[width=\textwidth]{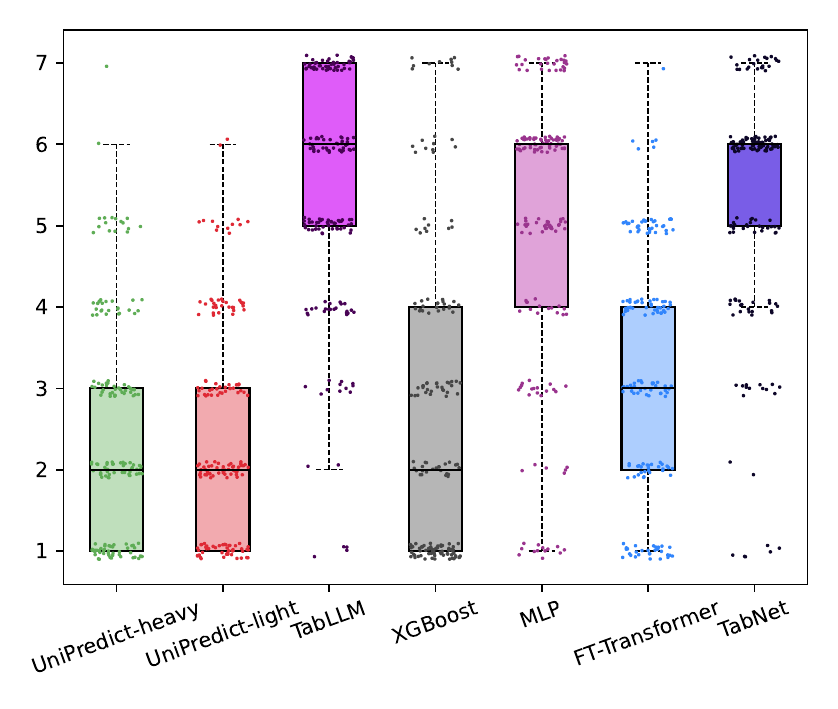}
         \caption{Model Rank}
         \label{fig:supervised-dataset-result-rank}
     \end{subfigure}
     \caption{The average accuracy and rank of \texttt{UniPredict-heavy}, \texttt{UniPredict-light}, \texttt{TabLLM} \citep{hegselmann2023tabllm} \texttt{XGBoost} \citep{chen2016xgboost}, \texttt{MLP}, \texttt{TabNet} \citep{arik2021tabnet} and \texttt{FT-Transformer} \citep{gorishniy2021revisiting} on 169 datasets. Each dot indicates a trial on one dataset. \texttt{UniPredict-heavy} demonstrates a remarkable performance advantage over the best neural network model (\texttt{FT-Transformer}) with a relative improvement of 13.4\%. It also surpasses the best-performing tree-boosting algorithms by a margin of 5.4\%.  Our framework's advantage is further confirmed by Figure \ref{fig:supervised-dataset-result-rank}, the model ranking (the less the better)}
    \label{fig:supervised-result}
    \vspace{-1em}
\end{figure*}

We assessed model accuracy on the test set of all 169 datasets and summarized the results in Figure~\ref{fig:supervised-result}. It is noted that due to the limitation of baseline models in terms of transferability onto new datasets, a distinct model was trained for each dataset, as discussed in Section \ref{sec:baseline}. Nonetheless, even without additional dataset-specific fine-tuning, both variants of \texttt{UniPredict} consistently outperform all baseline models in terms of accuracy.

Specifically, \texttt{UniPredict-heavy} achieves a notable increase in absolute accuracy of 2.2\% when compared to \texttt{XGBoost}, which stands as the top-performing model among the baseline models. Meanwhile, \texttt{UniPredict-light}, following in the footsteps of its full-size counterpart, continues to exhibit better performance relative to the other models. The ranking metric confirms their dominance over the baselines. In this metric, both \texttt{UniPredict-heavy} and \texttt{UniPredict-light} consistently occupy top positions.
As a candidate of tree-boosting method, although \texttt{XGBoost} shares a similar median ranking with the best-performing models, it displays a higher 25\% quartile in Figure~\ref{fig:supervised-dataset-result-rank}, indicating a sparser distribution of rankings.
The other baselines fail to deliver comparable performance. \texttt{TabLLM}, designed as an LLM-driven model for individual datasets, does not yield results that are on par with other lighter methods. Despite its moderate ranking in terms of accuracy, it falls to the lower ranks when considering median ranking. Further details on dataset-specific results regarding accuracy and rank are provided in Appendix \ref{apdx:train-result-all}.

\subsection{Results on Few-shot Learning}
\label{sec:few-shot-results}
We experimented \texttt{UniPredict}'s few-shot learning accuracy, compared with baseline models that are trained individually on each of the 62 datasets, where each dataset contains less than $100$ samples. 
This setup is to evaluate models on low-resource datasets because (1) collecting high-quality samples is of high cost in practice, and (2) models that generalize well in large datasets do not always perform as well as in small datasets. For each dataset, we divided it into a train set and a test set, which served for training each model and fine-tuning the pre-trained \texttt{UniPredict} and \texttt{TabLLM}. To thoroughly assess our model's capacity for generalization, we devised multiple experimental configurations involving the partitioning of the training dataset into different proportions, spanning from 10\% to 90\% of the entire dataset. For each of these settings, we trained separate baseline models on the respective datasets.

\begin{figure*}
     \centering
     \begin{subfigure}[b]{0.45\textwidth}
         \centering
         \includegraphics[width=\textwidth]{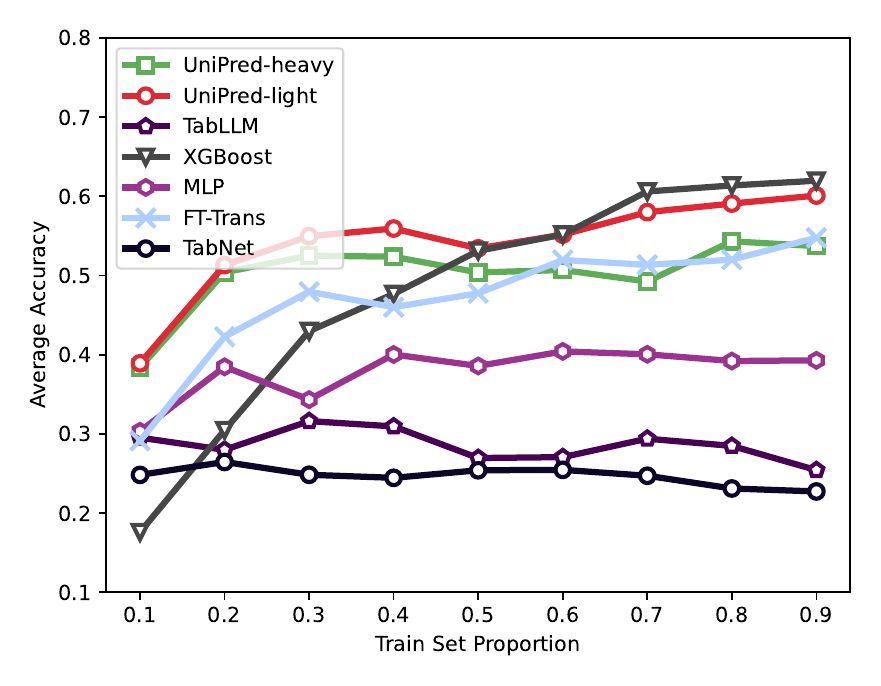}
         \caption{Model Accuracy (The higher the better)}
         \label{fig:fine-tuning-result-accuracy}
     \end{subfigure}
     \hfill
     \begin{subfigure}[b]{0.45\textwidth}
         \centering
         \includegraphics[width=\textwidth]{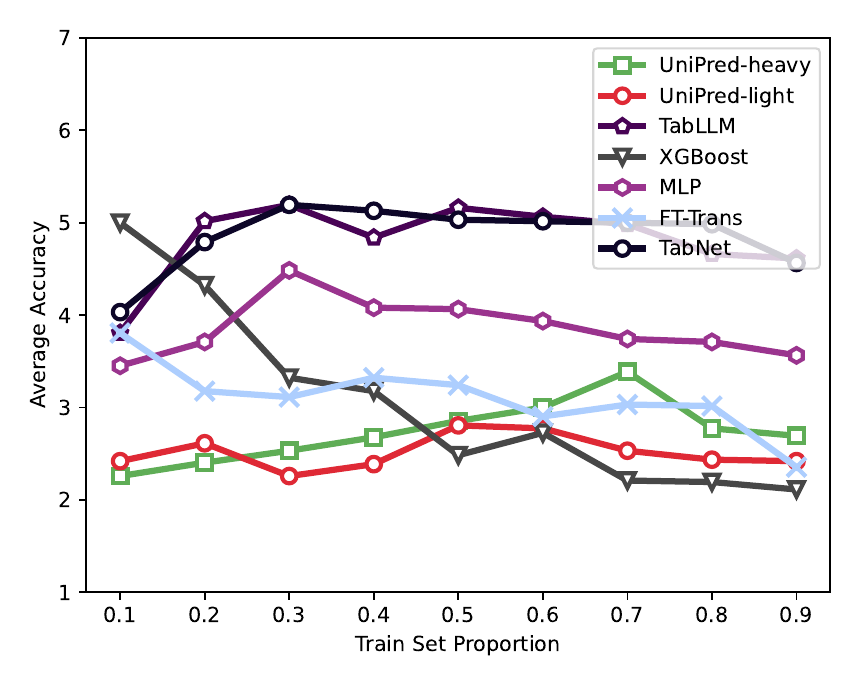}
         \caption{Model Rank (The lower the better)}
         \label{fig:fine-tuning-result-rank}
     \end{subfigure}
     \caption{The average accuracy and rank of \texttt{UniPredict-heavy}, \texttt{UniPredict-light}, \texttt{TabLLM} \texttt{XGBoost}, \texttt{MLP}, \texttt{TabNet} and \texttt{FT-Transformer} on 62 datasets.  We vary the training data size, ranging from the lowest (10\%) to the highest (90\%) of the full dataset. The pre-trained \method series exhibit remarkable data efficiency in generalizing to new tasks.}
    \label{fig:fine-tuning-result}
    \vspace{-1em}
\end{figure*}

Figure \ref{fig:fine-tuning-result} shows the accuracy and ranking of all models with varying training data sizes. The \texttt{UniPredict} series demonstrates a significant advantage in the low-resource regime, particularly when the training sets contain less than 50\% of the samples. As the sample size increases, they consistently remain among the top-performing models. The same trend is reflected in the result of model rankings as illustrated in Figure~\ref{fig:fine-tuning-result-rank}.
In contrast, \texttt{XGBoost} shines as the best model in resource-rich training setups, achieving an average accuracy of 0.62 when the training set size is set to 90\% of the entire dataset. However, it struggles in scenarios with small training sets. In the extreme low-resource case, where the training set proportion is 10\%, it exhibits the poorest performance among all models, with an over 118\% disadvantage to \texttt{UniPred-heavy}, and ranks at the bottom.
On the other hand, \texttt{FT-Transformer}, an attention-based model, performs comparably to \texttt{UniPredict-heavy} but falls short of surpassing either \texttt{UniPredict-light} or \texttt{XGBoost} in any of the setups. Its rank, however, jumped to the second in the last experiment setup on Figure \ref{fig:fine-tuning-result-rank}.
\texttt{MLP} delivers a moderate performance, while \texttt{TabNet} fails to converge effectively in these experimental setups. Similarly, \texttt{TabLLM} encounters problems in this context. Throughout all conditions, both \texttt{TabLLM} and \texttt{TabNet} consistently rank at the bottom and do not demonstrate improvement as the training set size scales up.
Additional information is provided in Appendix \ref{apdx:few-shot} for more detailed performance analysis of all models.

\subsection{Achilles' Heel: \texttt{UniPredict}'s failure analysis} \label{sec:case-study}
In this section, we aim to explore situations where our \method framework does not perform well, which provides insight for deploying \method and further enhancement. We have identified these situations by collecting datasets from the supervised setup (as used in Section \ref{sec:training-results}) and identifying the datasets in which either \texttt{UniPredict-heavy} or \texttt{UniPredict-light} ranks in the bottom 2 (6th or 7th) among all compared methods. For each of these datasets, we have collected potential causes that may lead to the poor performance of our method. We conclude that most failures can be attributed to one or more of the following causes:

\begin{itemize}[leftmargin=*]
\item \textbf{COL:} Too many \textbf{COL}umns in the dataset. This may result in serialized input strings that exceed the context limit of the language model. It hence undermines model performance because the exceeding parts are pruned.
\item \textbf{FV:} Poorly represented \textbf{F}eature \textbf{V}alues that are challenging for the model to process and comprehend. Examples include an excessive number of numerical values or meaningless characters.
\item \textbf{META:} Inadequate or ambiguous \textbf{META}Data, such as vague or meaningless column names and metadata, can confuse the model when comprehending the inputs.
\item \textbf{OTH:} \textbf{OTH}er factors not explicitly covered above that may deteriorate model performance.
\end{itemize}

\begin{figure*}
     \centering
     \begin{subfigure}[b]{0.3\textwidth}
         \centering
         \includegraphics[width=\textwidth]{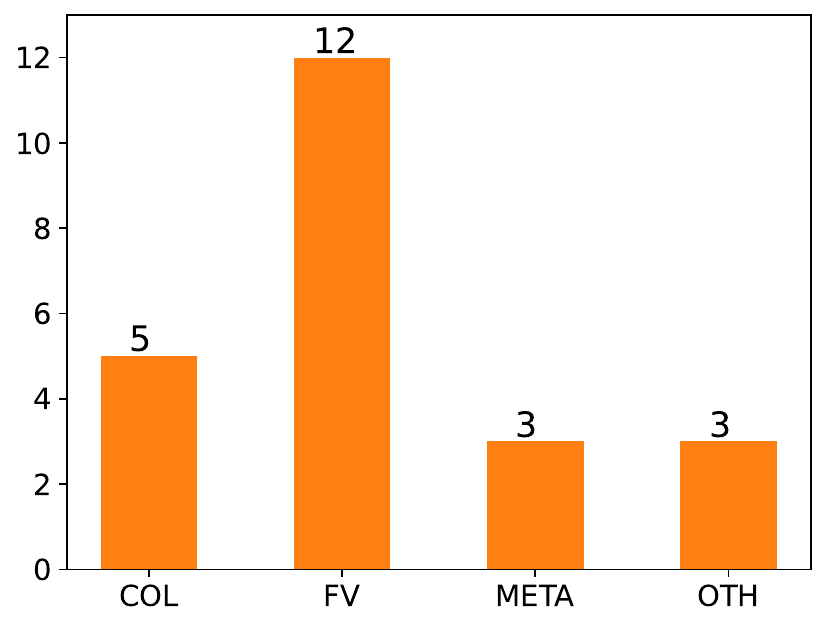}
         \caption{Overview}
         \label{fig:5a}
     \end{subfigure}
     \hfill
     \begin{subfigure}[b]{0.3\textwidth}
         \centering
         \includegraphics[width=\textwidth]{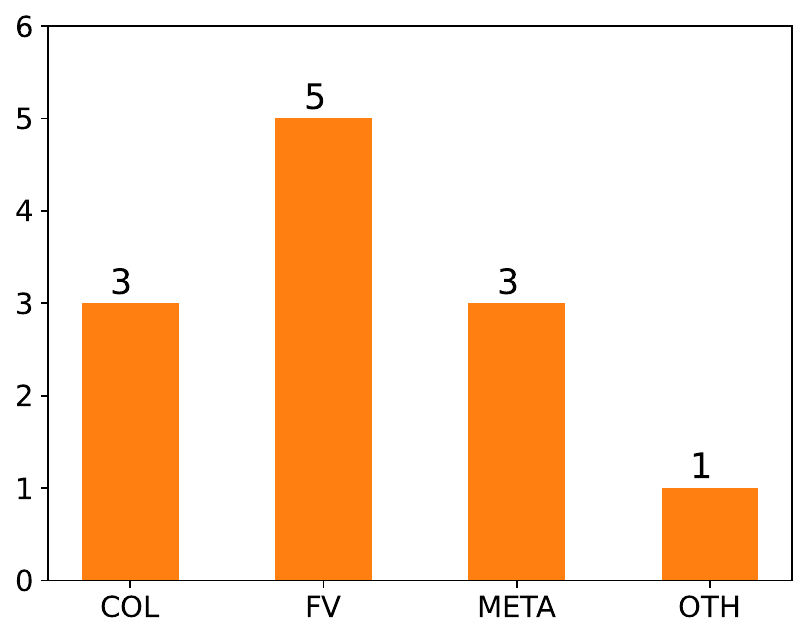}
         \caption{\texttt{UniPredict-heavy}}
         \label{fig:5b}
     \end{subfigure}
     \hfill
     \begin{subfigure}[b]{0.3\textwidth}
         \centering
         \includegraphics[width=\textwidth]{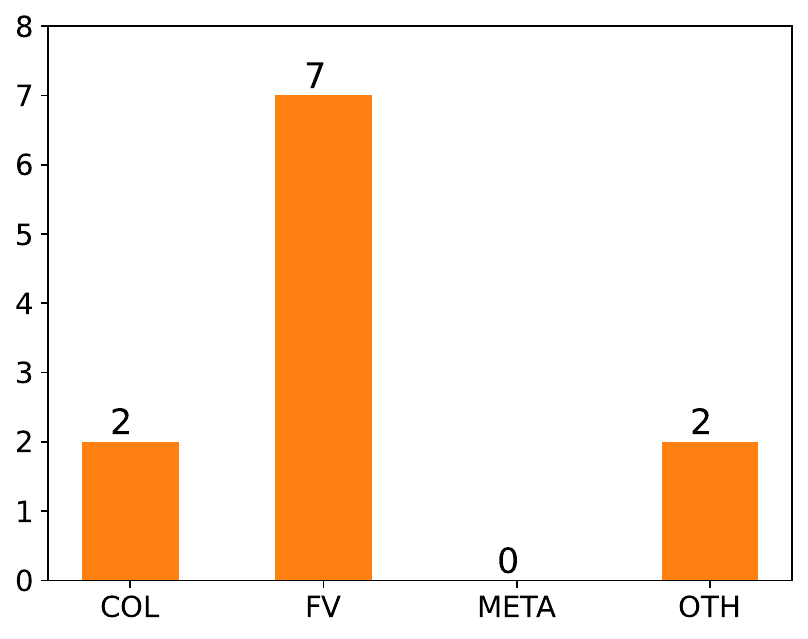}
         \caption{\texttt{UniPredict-light}}
         \label{fig:5c}
     \end{subfigure}
        \caption{an overview of the causes for which either model (Figure \ref{fig:5a}), \texttt{UniPredict-heavy} (Figure \ref{fig:5b}), or \texttt{UniPredict-light} (Figure \ref{fig:5c}) experienced poor performance. As described in Section \ref{sec:case-study},  \textbf{COL}, \textbf{FV}, \textbf{META} and \textbf{OTH} stand for  \textit{Excessive Column Number}, \textit{Bad Feature Values}, \textit{Bad Metadata} and \textit{Other reasons}, respectively. Among the 169 datasets examined, 8 datasets are included in \texttt{UniPredict-heavy}'s investigation, with 12 causes identified. \texttt{UniPredict-light} fails on 10 datasets, with 11 causes identified.}
        \label{fig:case-study}
        \vspace{-1em}
\end{figure*}

We include examples of each causes in Appendix \ref{apdx: failure}.
As illustrated in Figure~\ref{fig:case-study}, bad feature values are the primary cause behind approximately half of the failures observed in our framework. Additionally, \texttt{UniPredict-heavy} is affected by confusing metadata descriptions and oversized columns. Interestingly, \texttt{UniPredict-light}, which is configured with minimal metadata usage (as discussed in Section \ref{sec:implementation}), seems poised to minimize the influence of poor metadata. However, it paradoxically appears to struggle more with uninterpretable feature values, leading it to encounter more instances of poor performance compared to the default setup, \texttt{UniPredict-heavy}. 

In a nutshell, we conclude with three hints in developing \method in practice: (1) offering informative and accurate metadata for the input tabular dataset; (2) improving the context window limit of the LLM predictor to process more complicated inputs; and (3) cleaning up bad feature values before the training.

\subsection{Ablation Study}\label{sec:ablation}

\begin{table*}[t]
    \centering
    \begin{tabular}{l|cccc}
    \hline
          & \textbf{UniP-h} & \textbf{Abl-h} & \textbf{UniP-l} & \textbf{Abl-l} \\
          \hline
        \textbf{Universal Tabular Modeling} (avg.) & 0.721 & 0.686 & 0.740 & 0.575 \\
        \textbf{Universal Tabular Modeling} (med.) & 0.810 & 0.746 & 0.790 & 0.590 \\
        \textbf{Few-Shot Learning: Low-data} (avg.) & 0.525 & 0.483 & 0.513 & 0.349 \\
        \textbf{Few-Shot Learning: Low-data} (med.) & 0.521 & 0.474 & 0.500 & 0.289 \\
        \textbf{Few-Shot Learning: High-data} (avg.) & 0.543 & 0.545 & 0.590 & 0.321 \\
        \textbf{Few-Shot Learning: High-data} (med.) & 0.563 & 0.571 & 0.645 & 0.333 \\
            \hline
    \end{tabular}
    \caption{The result of ablation among \texttt{UniPredict-heavy} (\textbf{UniP-h}), \texttt{UniPredict-heavy} without target augmentation (\textbf{Abl-h}), \texttt{UniPredict-light} (\textbf{UniP-l}), \texttt{UniPredict-light} without target augmentation (\textbf{Abl-l}). Tasks examined are \textbf{Univeral Tabular Modeling} that uses the same set up as Section \ref{sec:training-results}, and \textbf{Few-shot Learning} as Section \ref{sec:few-shot-results}. The latter task involves both a low-data setup (Train Set Proportion = 0.3) and a high-data setup (Train Set Proportion = 0.8), which correspond to the conditions shown in Figure \ref{fig:fine-tuning-result}. For each task and setup, we provide both the \texttt{average} and \texttt{median} performance metrics across all datasets.}
    \label{tab:ablation}
\end{table*}

In this section, we conduct an ablation study to examine whether the re-formatting and augmenting of targets are the critical factors contributing to the success of \texttt{UniPredict}. The results are presented in Table~\ref{tab:ablation}. In the ablation study, the language models were fine-tuned using labels that only contained the one-hot encoding of the target class without the confidence information distributed into classes. The results consistently demonstrate that regardless of the model variant (whether \texttt{light} or \texttt{heavy}), the model with target augmentation performs noticeably better than the model without augmentation.
Furthermore, it is noteworthy that the ablation of \texttt{UniPredict-light} results in a more significant decrease in performance compared to \texttt{UniPredict-heavy}. This finding aligns with the conjecture made in Section~\ref{sec:implementation} that the \texttt{heavy} variant is more robust and adaptable across different implementations and scenarios.

\section{Related Work}
\label{sec:related-works}

\textbf{Tabular Prediction}. Tree-based models have shown outstanding performance on tabular prediction tasks \citep{chen2016xgboost,ke2017lightgbm}. Inspired by the rise of deep learning for tabular prediction \citep{arik2021tabnet}, the recent research has emphasized three ways of improvement: (1) taking advantage of pre-training or transfer learning on broad tabular data \citep{wang2022transtab,zhu2023xtab}; (2) adapting pre-trained large language models to generate the target label column as the prediction \citep{dinh2022lift,hegselmann2023tabllm}; and (3) mining the graph structure considering an overview of the tabular dataset \citep{du2022learning,chen2023hytrel}. In addition, \citet{wang2023anypredict} unify tabular data from various sources into a natural language format, establishing a tabular prediction pipeline capable of handling diverse inputs. However, most of these algorithms perform discriminative modeling for tabular prediction and hence are restricted to making the prediction for a fixed target. \method, by contrast, depends on generative modeling for the prediction of any user-specified target.

\textbf{Large Language Model}. LLMs have demonstrated remarkable capabilities in logical thinking and solving language tasks under instructions \citep{bubeck2023sparks, zhao2023survey}. It has motivated researchers to adopt LLMs for a series of tabular data tasks, including tabular data generation \citep{borisov2022language} and table-to-text generation \citep{zhao2023large}. Meanwhile, LLMs are fine-tuned for tabular prediction as generation task \citep{dinh2022lift,hegselmann2023tabllm}. While these studies have showcased LLM is able to generate target labels given textualized tabular data, there remains an unexplored opportunity: constructing a versatile tabular predictor capable of handling a wide array of tabular datasets. In addition, previous LLM-based tabular predictors are usually trained to generate the target label while not offering the corresponding prediction probabilities. We argue it is crucial to inspect the prediction probabilities made by LLMs, which is necessary when deploying them in production.

\section{Conclusion}\label{sec:conclusion}
We present \method that can learn from an aggregation of widespread tabular datasets called universal tabular prediction. We train a single \method model on 169 datasets with more than 300,000 samples and test it on the other 62 datasets for few-shot learning. Empirically, \texttt{UniPredict} yields the best prediction accuracy of 0.81 ($2.2\%$ absolute, $5.4\%$ relative improvement compared to \texttt{XGBoost}).
On unseen datasets, after dataset-specific fine-tuning, it exhibits great advantage when the training sets contain less than 50\% of the samples (118\% relative advantage to \texttt{XGBoost} at train-ratio=0.1) and consistently ranks at the top 2 in all scenarios. We envision that \method paves the way for deploying foundational tabular prediction systems.

\nocite{langley00}

\bibliography{example_paper}
\bibliographystyle{icml2024}

\newpage
\appendix
\onecolumn
\section{Methodology: More Detail}
\subsection{Prompt Templates} \label{apdx:prompt}
The quality of the natural language input provided to Large Language Models (LLMs) play a crucial role in determining the model's output and, consequently, its performance on tabular prediction tasks. The following are the prompt templates used in the implementation of both \texttt{UniPredict-heavy} and \texttt{UniPredict-light}:

\begin{lstlisting}[caption={Prompt for \texttt{UniPredict-heavy}}, language=python]
"""
    Below is the description of a dataset, an object profile from the dataset and a target description. Predict the target by the given information of the object.\n
    # Dataset description: {metadata}\n
    # Object description: {features}\n
    # You should return the probability of each class by: \n{instructions}\n
    # Answer: \n
"""
\end{lstlisting}

\begin{lstlisting}[caption={Prompt for \texttt{UniPredict-light}}, language=python]
"""
    Below is a dataset. Predict the target by the given information of the object.\n
    # Object description: {features}\n
    # You should return the probability of each class by: \n{instructions}\n
    # Answer: \n
"""
\end{lstlisting}
The key distinction between \texttt{UniPredict-heavy} and \texttt{UniPredict-light} lies in the utilization of re-formatted metadata information. \texttt{UniPredict-heavy} incorporates this re-formatted metadata to enhance the language model's understanding of the dataset context and schema, while \texttt{UniPredict-light} opts not to include this information to maintain a lighter and more concise prompting approach. We talk about the \textbf{metadata re-formatting} procedure in Section \ref{sec:metadata} and Appendix \ref{apdx:metadata}.

\subsection{Metadata Reformatting} \label{apdx:metadata}
Metadata often goes overlooked in data analysis as traditional models and algorithms do not typically incorporate them as part of the input. However, metadata can provide valuable insights and context for various aspects of data analysis, including the dataset's purpose and the target for prediction. In our framework, we actively collect metadata from two sources:
\begin{itemize}[leftmargin=*]
    \item \textbf{Dataset Descriptions}, which usually appear in the front page of the dataset as an introduction.
    \item \textbf{Column values}, which can be found inside of the datasheet.
\end{itemize}
With this information, we generate re-formatted dataset metadata for the following subjects:
\begin{itemize}[leftmargin=*]
    \item \textbf{Dataset Purpose} This section states the purpose of the dataset, providing necessary context and background information.
    \item \textbf{Target} This section specifies the item within the dataset that should be the target for prediction.
    \item \textbf{Column meanings} This section explains the meaning of columns, especially in cases where column names may not directly map to semantic meanings (e.g., columns labeled 'a', 'b', 'c', etc.). It also elaborates on the significance of each column, often drawing from the dataset description to provide a more comprehensive understanding.
\end{itemize}
In our implementation, we use the \texttt{gpt-3.5-turbo} model via the \textbf{OpenAI-API} to facilitate metadata re-formatting.
Our prompt input to \texttt{gpt-3.5} is shown as below:

\begin{lstlisting}[caption={Prompt for metadata re-formatting via \textbf{OpenAI-API}}, language=python]
"""
    The following is the metadata of a tabular dataset. Return the information for:\n
        1. the target of the dataset. If no target exists, choose one from the column as target for the dataset to classify.\n
        2. the features and their explanations, or N/A if there are no explanations. Replace all hyphens and/or underscores with spaces.\n\n
    Give your output in json. The following is an example output:\n
    '{\n'
    '    "target": "Age",\\n'
    '    "metadata": "The target of the dataset is Age. \\n Features and their explanations:\\n    gender: an animal\'s gender.\\n    weight: an animal\'s actual weight, in kg." \\n '
    '}\n\n'
    Do NOT respond anything else than the needed information. Make it brief but informative.
    Your responses should only be code, without explanation or formatting.\n\n
    columns:{col}\n\n
    metadata:{metadata}\n
    Provide your response in stringfied JSON format.
"""
\end{lstlisting}

Example inputs that are filled into this prompt template are as follows:
\begin{lstlisting}[caption={Example input to the prompt for metadata re-formatting. Information origin: \texttt{arnavsmayan-netflix-userbase-dataset}}, language=python]
metadata = "The dataset provides a snapshot of a sample Netflix userbase, showcasing various aspects of user subscriptions, revenue, account details, and activity. Each row represents a unique user, identified by their User ID. The dataset includes information such as the user's subscription type (Basic, Standard, or Premium), the monthly revenue generated from their subscription, the date they joined Netflix (Join Date), the date of their last payment (Last Payment Date), and the country in which they are located.\n\nAdditional columns have been included to provide insights into user behavior and preferences. These columns include Device Type (e.g., Smart TV, Mobile, Desktop, Tablet), Total Watch Time (in minutes), and Account Status (whether the account is active or not). The dataset serves as a synthetic representation and does not reflect actual Netflix user data. It can be used for analysis and modeling to understand user trends, preferences, and revenue generation within a hypothetical Netflix userbase."

col = "User ID,Subscription Type,Monthly Revenue,Join Date,Last Payment Date,Country,Age,Gender,Device,Plan Duration"

\end{lstlisting}

The following is our expected metadata after being re-formatted:
\begin{lstlisting}[caption={Example output from metadata re-formatting. Result generated from: \texttt{arnavsmayan-netflix-userbase-dataset}}, language=python]
"""
    The target of the dataset is Subscription Type. \n Features and their explanations:\n User ID: unique identifier for each user.\n Monthly Revenue: the amount of revenue generated from each user's subscription.\n Join Date: the date when each user joined Netflix.\n Last Payment Date: the date of the last payment made by each user.\n Country: the country in which each user is located.\n Age: the age of each user.\n Gender: the gender of each user.\n Device: the type of device used by each user.\n Plan Duration: the duration of each user's subscription plan.
"""
\end{lstlisting}

\subsection{Feature Serialization Example} \label{apdx:feature}
We present 3 sample feature serializations from different datasets below:
\begin{lstlisting}[caption={Feature serialization sample from \texttt{arnavsmayan-netflix-userbase-dataset}.}, language=python]
columns = "User ID,Subscription Type,Monthly Revenue,Join Date,Last Payment Date,Country,Age,Gender,Device,Plan Duration"
values = "1448,Standard,14,18-07-22,07-07-23,United States,33,Female,Laptop,1 Month"
# result: "'User ID is 1448; Monthly Revenue is 14; Join Date is 18-07-22; Last Payment Date is 07-07-23; Country is United States; Age is 33; Gender is Female; Device is Laptop; Plan Duration is 1 Month.\n'"
\end{lstlisting}
\begin{lstlisting}[caption={Feature serialization sample from \texttt{tarkkaanko-amazon}.}, language=python]
columns = ",reviewerName,overall,reviewText,reviewTime,day_diff,helpful_yes,helpful_no,total_vote,score_pos_neg_diff,score_average_rating,wilson_lower_bound"
values = "2346,J. Morse,5.0,'When I opened the micro disc and adapter I did't know what to do with them. I went to UTube on installing them, and all became clear. The micro fits into the top of the adapter and then the whole thing fits into my camera. Very neat and high powered.',2013-09-09,455,0,0,0,0,0.0,0.0"
# result: "Unnamed: 0 is 2346; reviewerName is J. Morse; reviewText is When I opened the micro disc and adapter I did't know what to do with them. I went to UTube on installing them, and all became clear. The micro fits into the top of the adapter and then the whole thing fits into my camera. Very neat and high powered.; reviewTime is 2013-09-09; day diff is 455; helpful yes is 0; helpful no is 0; total vote is 0; score pos neg diff is 0; score average rating is 0.0; wilson lower bound is 0.0.\n"
\end{lstlisting}
\begin{lstlisting}[caption={Feature serialization sample from \texttt{whenamancodes-predict-diabities}.}, language=python]
columns = "Pregnancies,Glucose,BloodPressure,SkinThickness,Insulin,BMI,DiabetesPedigreeFunction,Age,Outcome"
values = "6,98,58,33,190,34,0.43,43,0"
# result: 'Pregnancies is 6.0; Glucose is 98.0; BloodPressure is 58.0; SkinThickness is 33.0; Insulin is 190.0; BMI is 34.0; DiabetesPedigreeFunction is 0.43; Age is 43.0.\n'
\end{lstlisting}

\subsection{Target Augmentation}\label{apdx:target}
As explained in Section \ref{sec:target-augmentation}, we re-format the targets into one-hot encodings and assign probabilities to them rather than using the one-hot binary labels ($l \in {0, 1}$). The process of producing one-hot encodings depends on the nature of the target: If the targets are continuous values, we cluster the them into four quarters within the domain and represent them as categories; if the targets are already discrete values, we directly use the target value as the categories. The results are then serialized to be the reference output that the model is using for training.
We provide specific examples for each of these implementations below:
\begin{lstlisting}[caption={Descrete target augmentation example. Data come from \texttt{arnavsmayan-netflix-userbase-dataset}.}, language=python]
# target_space: ['Standard', 'Premium', 'Basic']
example_target = ['Premium']
target_after_one_hot = [0, 1, 0]
target_after_augmentation = [0.32, 0.39, 0.29]

# outcome from target augmentation:
target_class_details = 'class 0 stands for "Standard"; class 1 stands for "Premium"; class 2 stands for "Basic"'
target_serialization = 'class 0: 0.32; class 1: 0.39; class 2: 0.29.'

\end{lstlisting}
\begin{lstlisting}[caption={Continuous target augmentation example. Data come from \texttt{mirichoi0218-insurance}.}, language=python]
# target_space = 1121 - 63770
# categorized_target_space: ["<4740.0", "4740.0 - 9380.0", "9380.0 - 16600.0", ">16600.0"]
example_target = ['9095.069']

# outcome from target augmentation:
target_class_details = 'class 0 stands for ">16600.0"; class 1 stands for "<4740.0"; class 2 stands for "9380.0 - 16600.0"'
target_serialization = 'class 0: 0.09; class 1: 0.0; class 2: 0.05; class 3: 0.86.'
\end{lstlisting}

\subsection{LLM Output Mapping}\label{apdx:output-mapping}
For an LLM output that follows the format we described in Section \ref{apdx:target}, we can use \texttt{Regex} matching to capture model's prediction. 

Let 
\begin{lstlisting}[language=python]
response = 'class 0: 0.09; class 1: 0.0; class 2: 0.05; class 3: 0.86.'
\end{lstlisting}
be a sample response from the LLM, we obtain a listed result of numerical probabilities by applying
\begin{lstlisting}[language=python]
result = re.findall(r'[\d]*[.][\d]+', response)
# result = [0.09, 0.0, 0.05, 0.86]
\end{lstlisting}
Based on the listed result, we can compute the model's prediction on classes by finding the index of the maximum in the list. 
\begin{lstlisting}[language=python]
result_class = pred_cls.index(max(result))
# result_class = 3
\end{lstlisting}
\section{Implementation Details}
\subsection{Baseline}\label{apdx:baseline}
In this section, we present our baseline setups:
\begin{itemize}[leftmargin=*]
    \item \textbf{XGBoost} is a tree-ensemble method that has been broadly used in tabular prediction. In our experiment, we train \texttt{XGBoost} instances via its official release on Python. \footnote{Information can be found at https://xgboost.readthedocs.io/en/stable/python/python\_intro.html.} We apply ordinal encoding on all features and categories except the numerical features and tune one instance on each dataset with \texttt{n\_estimators=100, max\_depth=6, learning\_rate=0.3}.
    \item \textbf{Multilayer Perceptron} is a fundamental neural network architecture that consists of fully-connected hidden layers. We use the \texttt{MLPClassifier} instance from \texttt{scikit-learn}. On each dataset, a model is instantiated with \texttt{learning\_rate=1e-3, n\_hidden\_layer=1, activation='relu', optimizer='adam'}. We also set \texttt{random\_state=1} and \texttt{max\_iteration=100}.
    \item \textbf{FT-Transformer} is an attention-based model designed and trained specifically for tabular data tasks. We use the original implementation from the author \footnote{https://github.com/Yura52/rtdl} with no extra changes on implementation. The hyperparameters we use here are \texttt{num\_batchs=8, num\_epochs=100, learning\_rate=1e-3}.
    \item \textbf{TabNet} is another attention based model on tabular data. We instantiate models from its official release on python \footnote{https://pypi.org/project/pytorch-tabnet/}. Similar to our approach with \texttt{XGBoost}, we applied the same data preprocessing procedure to \texttt{TabNet}. Specifically, we used ordinal encoding for features and categories (excluding numerical features). We conducted model tuning using the default hyperparameters.
    \item \textbf{TabLLM} is an LLM-based system specifically designed for tabular prediction tasks. In our implementation, we followed the setup as described in the original work. Since \method is built upon a GPT-2 backbone, we implement TabLLM on a GPT-2 as well to align the backbone choices for a fair comparison. When incorporating specific instructions into the prompt, instead of creating separate instances to ask 'yes-or-no' questions individually for each target class, we streamlined the process by instructing the model to predict the class name directly. This approach simplifies the training procedure and conserves computational resources. An example prompt is presented below. We train isolated \texttt{TabLLM} instances on each dataset, regardless of the origin of the dataset (supervised division or few-shot division).
\end{itemize}

\begin{lstlisting}[caption={Prompt for \texttt{TabLLM}}, language=python]
"""
    Below is a dataset. Predict the target by the given information of the object.\n
    # Object description: {features}\n
    # You should return your choice of class by stating the class number, {instructions}\n
    # Answer: \n
"""
# 'instructions' includes a sequence stating the detail of each class, for example 'class 1 is for "a", class 2 is for "b", ...'
# Example model output: 'class 1'
\end{lstlisting}

\subsection{Dataset Statistics}\label{apdx:dataset-statistics}
We present all dataset statistics in Table \ref{tab:supervised-dataset-stats} and Table \ref{tab:few-shot-dataset-stats}. In the training setup, all datasets are split with a train-set-ratio=$0.9$. In the few-shot testing setup, all datasets are tested with different train set ratios ranging from $0.1$ to $0.9$.

\fontsize{8}{11}\selectfont
\setlength{\tabcolsep}{3pt}
\begin{longtabu} to \textwidth{p{2in} c c c | p{2in} c c c}
\caption{Dataset statistics for model training and testing (Results shown in Section \ref{sec:training-results}). We include each dataset's \textbf{Name}, number of \textbf{rows}, number of \textbf{cols}, and whether the dataset's targets are continuous (\textbf{Ctns}). The last measurement determines whether the dataset's targets need to be re-categorized into quarters, as detailed in Appendix \ref{apdx:target}.} \\
\textbf{Name} &  \textbf{rows} & \textbf{cols} & \textbf{Ctns} & \textbf{Name} &  \textbf{rows} & \textbf{cols} & \textbf{Ctns} \\
\hline        
arnavsmayan-netflix-userbase-dataset & 2500 & 9 & False & 
deependraverma13-diabetes-healthcare-comprehensive-dataset & 768 & 8 & False\\
bhanupratapbiswas-uber-data-analysis & 1156 & 6 & False & 
swathiunnikrishnan-amazon-consumer-behaviour-dataset & 602 & 22 & False\\
hemanthhari-psycological-effects-of-covid & 1175 & 21 & False & 
arslanr369-bitcoin-price-2014-2023 & 3228 & 6 & False\\
saloni1712-chatgpt-app-reviews & 2292 & 3 & True & 
naveenkumar20bps1137-predict-students-dropout-and-academic-success & 4424 & 34 & False\\
sanjanchaudhari-user-behavior-on-instagram & 7488 & 8 & False & 
bhanupratapbiswas-bollywood-actress-name-and-movie-list & 1284 & 9 & False\\
arnavsmayan-vehicle-manufacturing-dataset & 2000 & 7 & False & 
bharath011-heart-disease-classification-dataset & 1319 & 8 & False\\
shroukgomaa-babies-food-ingredients & 696 & 25 & False & 
amirhosseinmirzaie-countries-life-expectancy & 2848 & 17 & False\\
amirhosseinmirzaie-pistachio-types-detection & 1718 & 16 & False & 
shashankshukla123123-marketing-campaign & 2240 & 29 & False\\
uciml-pima-indians-diabetes-database & 768 & 8 & False & 
shubhamgupta012-titanic-dataset & 889 & 8 & False\\
bhanupratapbiswas-fashion-products & 1000 & 8 & False & 
blastchar-telco-customer-churn & 7043 & 20 & False\\
mirichoi0218-insurance & 1338 & 6 & False & 
suraj520-dairy-goods-sales-dataset & 4325 & 22 & False\\
uciml-red-wine-quality-cortez-et-al-2009 & 1599 & 11 & False & 
akshaydattatraykhare-diabetes-dataset & 768 & 8 & False\\
arnabchaki-data-science-salaries-2023 & 3755 & 10 & False & 
prkhrawsthi-bitcoin-usd-daily-price-with-volume-2015-2023 & 3104 & 6 & False\\
hawkingcr-airbnb-for-boston-with-fraud-detection & 3585 & 20 & False & 
saunakghosh-nba-players-dataset & 5130 & 7 & False\\
rtatman-chocolate-bar-ratings & 1795 & 8 & False & 
pavansubhasht-ibm-hr-analytics-attrition-dataset & 1470 & 34 & False\\
gyanprakashkushwaha-laptop-price-prediction-cleaned-dataset & 1273 & 12 & False & 
fedesoriano-stroke-prediction-dataset & 5110 & 11 & False\\
bhanupratapbiswas-world-top-billionaires & 2614 & 21 & False & 
vstacknocopyright-blood-transfusion-service-center-data & 748 & 5 & False\\
ashishkumarjayswal-movies-updated-data & 4000 & 14 & False & 
bhanupratapbiswas-ipl-dataset-2008-2016 & 577 & 15 & False\\
mathchi-diabetes-data-set & 768 & 8 & False & 
harishkumardatalab-medical-insurance-price-prediction & 2772 & 6 & False\\
arslanr369-roblox-stock-pricing-2021-2023 & 572 & 6 & False & 
yasserh-titanic-dataset & 891 & 11 & False\\
iqmansingh-company-employee-dataset & 5000 & 12 & False & 
shivamb-disney-movies-and-tv-shows & 1450 & 11 & False\\
alexisbcook-pakistan-intellectual-capital & 1142 & 12 & False & 
tahzeer-indian-startups-by-state & 7091 & 5 & False\\
harshitshankhdhar-imdb-dataset-of-top-1000-movies-and-tv-shows & 1000 & 15 & False & 
shreyapurohit-anime-data & 6850 & 4 & False\\
raddar-icr-integer-data & 617 & 57 & False & 
uciml-mushroom-classification & 8124 & 22 & False\\
adityakadiwal-water-potability & 3276 & 9 & False & 
shreyanshverma27-imdb-horror-chilling-movie-dataset & 836 & 7 & False\\
ruchi798-data-science-job-salaries & 607 & 11 & False & 
hesh97-titanicdataset-traincsv & 891 & 11 & False\\
phangud-spamcsv & 5572 & 1 & False & 
dileep070-heart-disease-prediction-using-logistic-regression & 4238 & 15 & False\\
abcsds-pokemon & 800 & 12 & False & 
atharvaingle-crop-recommendation-dataset & 2200 & 7 & False\\
rounakbanik-pokemon & 801 & 40 & False & 
thedevastator-cancer-patients-and-air-pollution-a-new-link & 1000 & 25 & False\\
andrewmvd-fetal-health-classification & 2126 & 21 & False & 
saurabh00007-diabetescsv & 768 & 8 & False\\
larsen0966-student-performance-data-set & 649 & 32 & False & 
nikhil1e9-netflix-stock-price & 5325 & 6 & False\\
yasserh-wine-quality-dataset & 1143 & 12 & False & 
ashishkumarjayswal-loanamount-approval & 614 & 12 & False\\
ananthr1-weather-prediction & 1461 & 5 & True & 
thedevastator-higher-education-predictors-of-student-retention & 4424 & 34 & False\\
rpaguirre-tesla-stock-price & 1692 & 6 & False & 
muhammadtsabitulazmi-liga-1-indonesia-player-dataset & 568 & 11 & False\\
ashishkumarjayswal-diabetes-dataset & 768 & 8 & False & 
wearefuture01-hepatitis-c-prediction & 615 & 13 & True\\
aakashjoshi123-exercise-and-fitness-metrics-dataset & 3864 & 11 & False & 
kumargh-pimaindiansdiabetescsv & 767 & 8 & False\\
gauravduttakiit-resume-dataset & 962 & 1 & False & 
surajjha101-stores-area-and-sales-data & 896 & 4 & False\\
rishikeshkonapure-hr-analytics-prediction & 1470 & 34 & False & 
eishkaran-heart-disease & 1190 & 11 & False\\
vikramamin-customer-churn-decision-tree-and-random-forest & 7043 & 20 & False & 
redwankarimsony-heart-disease-data & 920 & 15 & True\\
hashemi221022-diabetes & 768 & 8 & False & 
rajyellow46-wine-quality & 6497 & 12 & False\\
vikramamin-time-series-forecasting-using-prophet-in-r & 1827 & 4 & False & 
reihanenamdari-breast-cancer & 4024 & 15 & False\\
uciml-indian-liver-patient-records & 583 & 10 & False & 
teertha-ushealthinsurancedataset & 1338 & 6 & False\\
ninzaami-loan-predication & 614 & 12 & False & 
timoboz-tesla-stock-data-from-2010-to-2020 & 2416 & 6 & False\\
elakiricoder-gender-classification-dataset & 5001 & 7 & False & 
jainilcoder-netflix-stock-price-prediction & 1009 & 6 & False\\
burak3ergun-loan-data-set & 614 & 12 & False & 
sanjanchaudhari-bankloan & 1500 & 11 & False\\
alirezachahardoli-bank-personal-loan-1 & 5000 & 13 & False & 
sbhatti-financial-sentiment-analysis & 5842 & 1 & False\\
altruistdelhite04-gold-price-data & 2290 & 5 & False & 
carolzhangdc-imdb-5000-movie-dataset & 5043 & 27 & False\\
desalegngeb-german-fintech-companies & 978 & 23 & False & 
crxxom-manhwa-dataset & 2943 & 14 & False\\
varpit94-tesla-stock-data-updated-till-28jun2021 & 2956 & 6 & False & 
hashemi221022-bank-loans & 5000 & 13 & False\\
geomack-spotifyclassification & 2017 & 16 & False & 
jillanisofttech-brain-stroke-dataset & 4981 & 10 & False\\
mayankpatel14-second-hand-used-cars-data-set-linear-regression & 1000 & 11 & False & 
rkiattisak-student-performance-in-mathematics & 1000 & 7 & False\\
sabasaeed1953-stock-prices-of-2023 & 700 & 7 & False & 
primaryobjects-voicegender & 3168 & 20 & False\\
maryammanoochehry-bank-personal-loan & 5000 & 13 & False & 
bhavkaur-simplified-titanic-dataset & 2240 & 3 & False\\
sidhus-crab-age-prediction & 3893 & 8 & False & 
ahsan81-superstore-marketing-campaign-dataset & 2240 & 21 & False\\
fedesoriano-hepatitis-c-dataset & 615 & 13 & True & 
oles04-bundesliga-seasons & 5508 & 22 & False\\
gabrielsantello-cars-purchase-decision-dataset & 1000 & 4 & False & 
andrewmvd-udemy-courses & 3678 & 11 & False\\
whenamancodes-students-performance-in-exams & 1000 & 7 & False & 
patelprashant-employee-attrition & 1470 & 34 & False\\
barun2104-telecom-churn & 3333 & 10 & False & 
kandij-diabetes-dataset & 768 & 8 & False\\
vedavyasv-usa-housing & 5000 & 6 & False & 
team-ai-spam-text-message-classification & 5572 & 1 & False\\
prevek18-ames-housing-dataset & 2930 & 81 & False & 
mazlumi-ielts-writing-scored-essays-dataset & 1435 & 8 & False\\
vijayvvenkitesh-microsoft-stock-time-series-analysis & 1511 & 5 & False & 
ruchi798-tv-shows-on-netflix-prime-video-hulu-and-disney & 5368 & 11 & False\\
tarkkaanko-amazon & 4915 & 11 & True & 
kingabzpro-cosmetics-datasets & 1472 & 10 & False\\
receplyasolu-6k-weather-labeled-spotify-songs & 6368 & 5 & False & 
kabure-german-credit-data-with-risk & 1000 & 10 & False\\
mahnazarjmand-bank-personal-loan & 5000 & 13 & False & 
sudarshan6561-ipl-2023 & 568 & 4 & False\\
agirlcoding-all-space-missions-from-1957 & 4324 & 8 & False & 
mfaisalqureshi-spam-email & 5572 & 1 & False\\
cpluzshrijayan-milkquality & 1059 & 7 & False & 
awaiskaggler-insurance-csv & 1338 & 6 & False\\
thedevastator-employee-attrition-and-factors & 1470 & 34 & False & 
surajjha101-top-youtube-channels-data & 1000 & 6 & False\\
hansrobertson-american-companies-profits-and-benefits-from-ai & 1447 & 3 & False & 
dansbecker-aer-credit-card-data & 1319 & 11 & False\\
whenamancodes-predict-diabities & 768 & 8 & False & 
nancyalaswad90-review & 768 & 8 & False\\
ruchi798-student-feedback-survey-responses & 1001 & 9 & False & 
siddharthss-crop-recommendation-dataset & 2200 & 7 & False\\
therealsampat-predict-movie-success-rate & 839 & 32 & False & 
maryalebron-life-expectancy-data & 2938 & 24 & False\\
noordeen-insurance-premium-prediction & 1338 & 6 & False & 
ybifoundation-food-app-business & 2205 & 26 & False\\
oles04-top-leagues-player & 2612 & 17 & False & 
buntyshah-auto-insurance-claims-data & 1000 & 39 & False\\
lightonkalumba-us-womens-labor-force-participation & 753 & 22 & False & 
tejashvi14-employee-future-prediction & 4653 & 8 & False\\
arnabchaki-indian-restaurants-2023 & 6593 & 7 & False & 
kanths028-usa-housing & 5000 & 6 & False\\
ravibarnawal-mutual-funds-india-detailed & 814 & 19 & False & 
dsfelix-us-stores-sales & 4248 & 19 & False\\
sanjanchaudhari-netflix-dataset & 1818 & 10 & False & 
tejashvi14-engineering-placements-prediction & 2966 & 7 & False\\
bhavkaur-hotel-guests-dataset & 2000 & 9 & False & 
warcoder-earthquake-dataset & 782 & 18 & False\\
mayurdalvi-simple-linear-regression-placement-data & 1000 & 2 & False & 
arashnic-time-series-forecasting-with-yahoo-stock-price & 1825 & 6 & False\\
bretmathyer-telemedicine-used & 3344 & 14 & False & 
iamsumat-spotify-top-2000s-mega-dataset & 1994 & 14 & False\\
ahsan81-food-ordering-and-delivery-app-dataset & 1898 & 8 & False & 
kreeshrajani-human-stress-prediction & 2838 & 6 & False\\
shivamb-hm-stores-dataset & 4292 & 20 & True & 
christinestevens-cstevens-peloton-data & 3737 & 20 & False\\
aakashjoshi123-spotify-top-hits-data & 1000 & 6 & False & 
ishadss-productivity-prediction-of-garment-employees & 1197 & 14 & False\\
chirin-africa-economic-banking-and-systemic-crisis-data & 1059 & 13 & False & 
mayuriawati-bangalore-chain-restaurants-ratings-and-reviews & 1826 & 7 & False\\
azminetoushikwasi--lionel-messi-all-club-goals & 704 & 12 & False & &&&\\
\label{tab:supervised-dataset-stats}
\end{longtabu}

\begin{longtabu} to \textwidth{p{2in} c c c | p{2in} c c c}
\caption{Dataset statistics for the few-shot testing (Results shown in Section \ref{sec:few-shot-results}). We include each dataset's \textbf{Name}, number of \textbf{rows}, number of \textbf{cols}, and whether the dataset's targets are continuous (\textbf{Ctns}). The last measurement determines whether the dataset's targets need to be re-categorized into quarters, as detailed in Appendix \ref{apdx:target}.} \\
\textbf{Name} &  \textbf{rows} & \textbf{cols} & \textbf{Ctns} & \textbf{Name} &  \textbf{rows} & \textbf{cols} & \textbf{Ctns} \\
\hline        
mauryansshivam-paytm-revenue-users-transactions & 12 & 20 & False & 
yapwh1208-students-score & 56 & 12 & False\\
kagankoral-dceu-box-office-and-rating-dataset & 13 & 9 & False & 
drahulsingh-rohit-sharma-all-international-cricket-centuries & 43 & 8 & False\\
tapakah68-email-spam-classification & 84 & 2 & False & 
drahulsingh-s-chanderpaul-all-international-cricket-centuries & 41 & 8 & False\\
whydhruv-viratkohli-76centuries & 76 & 13 & False & 
drahulsingh-largest-banks & 100 & 3 & False\\
hammadjavaid-100-most-expensive-footballers-of-all-time & 101 & 8 & True & 
sanjanchaudhari-scheme-wise-placement-pmkvy & 18 & 7 & False\\
bhanupratapbiswas-national-youth-volunteers-2022-2023 & 37 & 11 & False & 
drahulsingh-top-largest-universities & 84 & 7 & False\\
drahulsingh-kane-williamson-all-cricket-centuries & 72 & 8 & False & 
abhijitdahatonde-india-population-1947-2011 & 37 & 8 & False\\
ravivarmaodugu-data-on-investment-and-employment-in-india & 49 & 4 & False & 
drahulsingh-mohammad-yousuf-all-cricket-centuries & 39 & 10 & False\\
abhishek14398-salary-dataset-simple-linear-regression & 30 & 2 & False & 
mauryansshivam-youtube-ads-revenue & 17 & 1 & False\\
sanjanchaudhari-pixarmovies & 15 & 15 & False & 
amirmotefaker-supply-chain-dataset & 100 & 23 & False\\
allanwandia-supply-chain-data & 31 & 22 & False & 
omarsobhy14-student-loans & 57 & 5 & False\\
drahulsingh-virat-kohli-all-international-cricket-centuries & 134 & 8 & False & 
hammadjavaid-highest-grossing-indian-movies-2023 & 105 & 8 & False\\
christph-harry-potter-potion-recipes & 132 & 3 & False & 
karthickveerakumar-salary-data-simple-linear-regression & 30 & 1 & False\\
sujithmandala-obesity-classification-dataset & 108 & 6 & False & 
harshsingh2209-supply-chain-analysis & 100 & 23 & False\\
drahulsingh-ross-taylor-all-international-cricket-centuries & 40 & 8 & False & 
sanjanchaudhari-us-employment-and-unemployment & 71 & 11 & False\\
anirudhkulkarni455-vande-bharat & 26 & 15 & False & 
yasserh-student-marks-dataset & 100 & 2 & False\\
dev523-cbse-class-x-result-data & 48 & 7 & False & 
drahulsingh-matthew-hayden-all-international-cricket-centuries & 40 & 8 & False\\
arindambaruah-void-formation-process-data-in-welding & 196 & 13 & False & 
ravitejakotharu-salary-datacsv & 30 & 1 & False\\
drahulsingh-chris-gayle-all-international-cricket-centuries & 42 & 9 & False & 
abhijitdahatonde-rohit-sharma-centuries & 43 & 10 & False\\
drahulsingh-hashim-amla-all-international-cricket-centuries & 55 & 8 & False & 
rsadiq-salary & 35 & 1 & False\\
codebreaker619-salary-data-with-age-and-experience & 30 & 2 & False & 
drahulsingh-ab-de-villiers-all-international-cricket-centuries & 47 & 8 & True\\
yusufdede-lung-cancer-dataset & 59 & 6 & False & 
mauryansshivam-netflix-ott-revenue-and-subscribers-csv-file & 17 & 14 & False\\
rohankayan-years-of-experience-and-salary-dataset & 30 & 1 & False & 
thamersekhri-liverpool-matches-dataset-2022-2023 & 59 & 39 & False\\
whenamancodes-impacts-of-energy-production & 14 & 22 & False & 
devchauhan1-salary-datacsv & 30 & 1 & False\\
ruiromanini-mtcars & 32 & 11 & False & 
maraglobosky-hot-dog-eating-contest-results & 62 & 7 & False\\
komalkhetlani-apple-iphone-data & 62 & 10 & False & 
anandhuh-latest-covid19-india-statewise-data & 36 & 8 & False\\
mathurinache-electriccarsalesbymodelinusa & 57 & 98 & False & 
fredericobreno-play-tennis & 14 & 5 & False\\
farhanmd29-50-startups & 50 & 4 & False & 
aaditshukla-beach-water-and-weather-sensor-locations & 9 & 4 & False\\
hussainnasirkhan-multiple-linear-regression-dataset & 20 & 2 & False & 
hb20007-gender-classification & 66 & 4 & False\\
usharengaraju-coursera-ipo-tweets & 8 & 35 & False & 
yashmerchant-cities & 73 & 5 & False\\
drahulsingh-rahul-dravid-all-international-cricket-centuries & 48 & 8 & False & 
fivethirtyeight-the-ultimate-halloween-candy-power-ranking & 85 & 12 & False\\

\label{tab:few-shot-dataset-stats}
\end{longtabu}

\fontsize{9.5}{8}\selectfont

\subsection{Model Training} \label{apdx:training}
We utilize a GPT-2 \citep{radford2019language} model as backbone. We perform training following an instruction fine-tuning process. The optimizer choice is \texttt{AdamW} with \texttt{lr=5e-5}, \texttt{beta\_1 = 0.9}, \texttt{beta\_2 = 0.999}, \texttt{epsilon = 1e-8}, \texttt{weight\_decay = 0}. The model is trained for \texttt{3} epochs. The model takes approximately 75 hours to be trained on a single RTX3090.

The few-shot learning process is almost identical to the training process described above. The only difference is that we increase the epoch to 30 to ensure convergence.

\section{Result}
\subsection{Detailed Model Performance on Universal Tabular Prediction}\label{apdx:train-result-all}
We present all models' performance on each supervised dataset in Table \ref{tab:supervised-results}, including the ablation models.

\subsection{Model Performance on Few-Shot Datasets}\label{apdx:few-shot}
\begin{figure}
     \centering
     \begin{subfigure}[b]{0.45\textwidth}
         \centering
         \includegraphics[width=\textwidth]{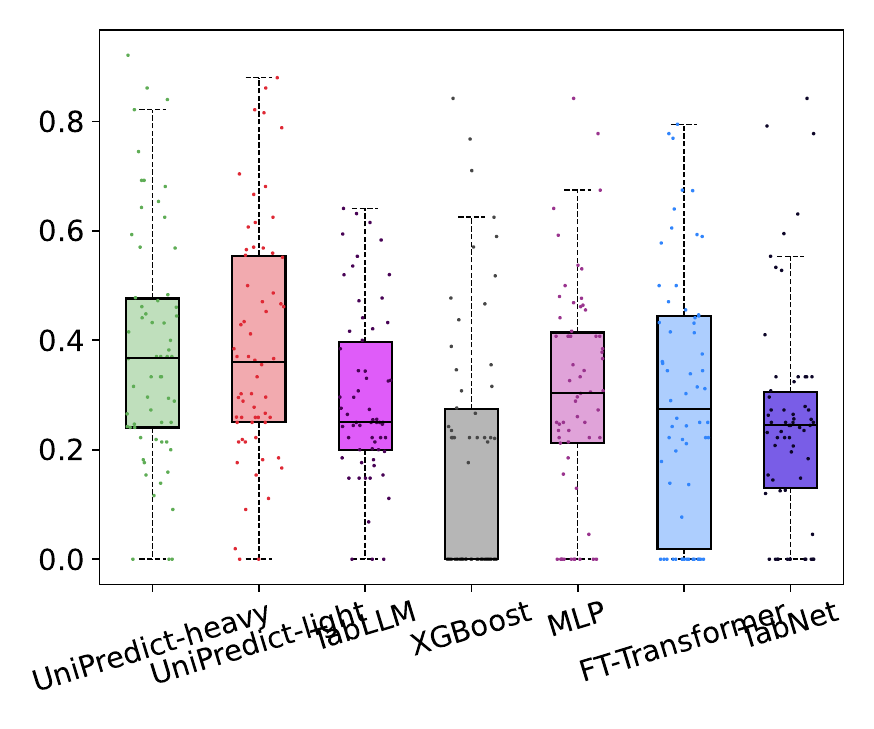}
         \caption{Model Accuracy (The higher the better)}
         \label{fig:few-shot-result-low-accuracy}
     \end{subfigure}
     \hfill
     \begin{subfigure}[b]{0.45\textwidth}
         \centering
         \includegraphics[width=\textwidth]{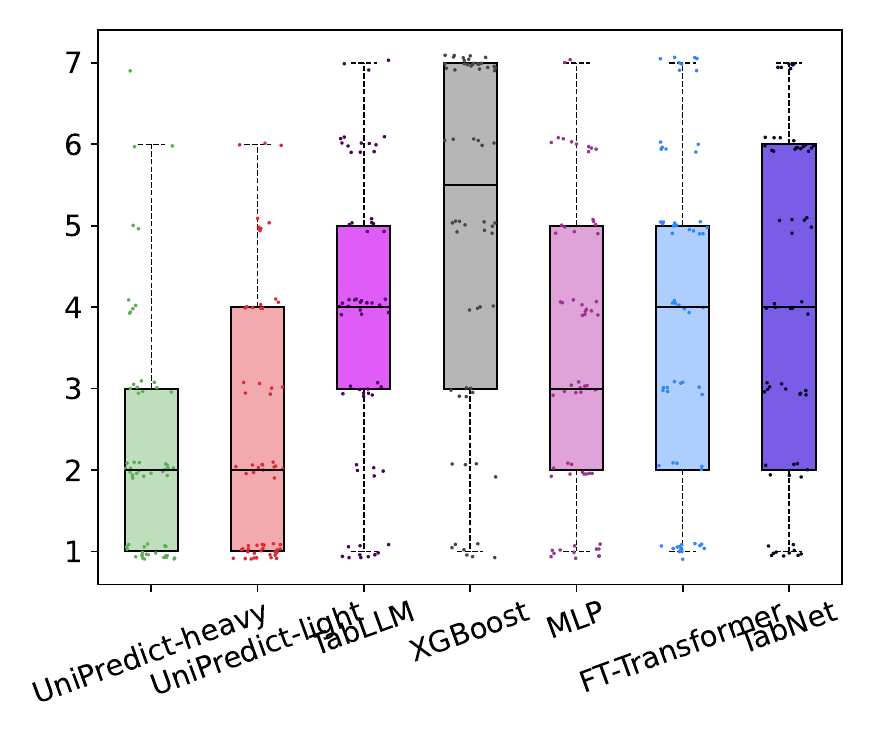}
         \caption{Model Rank (The lower the better)}
         \label{fig:few-shot-result-low-rank}
     \end{subfigure}
     \caption{The average accuracy and rank of \texttt{UniPredict-heavy}, \texttt{UniPredict-light}, \texttt{TabLLM} \texttt{XGBoost}, \texttt{MLP}, \texttt{TabNet} and \texttt{FT-Transformer} on the few-shot dataset with train-set-ratio set to 0.1.}
    \label{fig:few-shot-result-low}
\end{figure}

\begin{figure}
     \centering
     \begin{subfigure}[b]{0.45\textwidth}
         \centering
         \includegraphics[width=\textwidth]{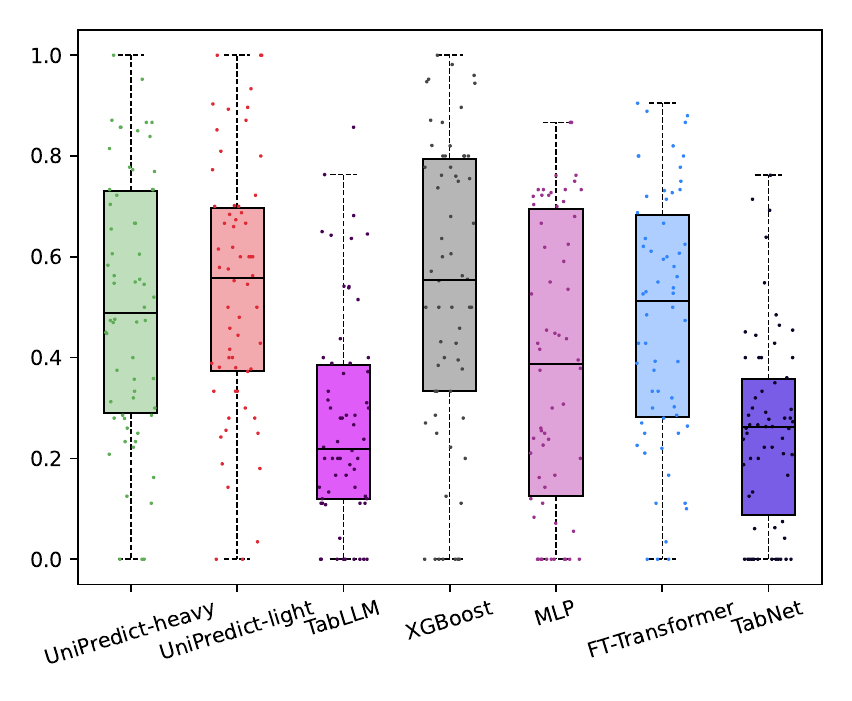}
         \caption{Model Accuracy (The higher the better)}
         \label{fig:few-shot-result-mid-accuracy}
     \end{subfigure}
     \hfill
     \begin{subfigure}[b]{0.45\textwidth}
         \centering
         \includegraphics[width=\textwidth]{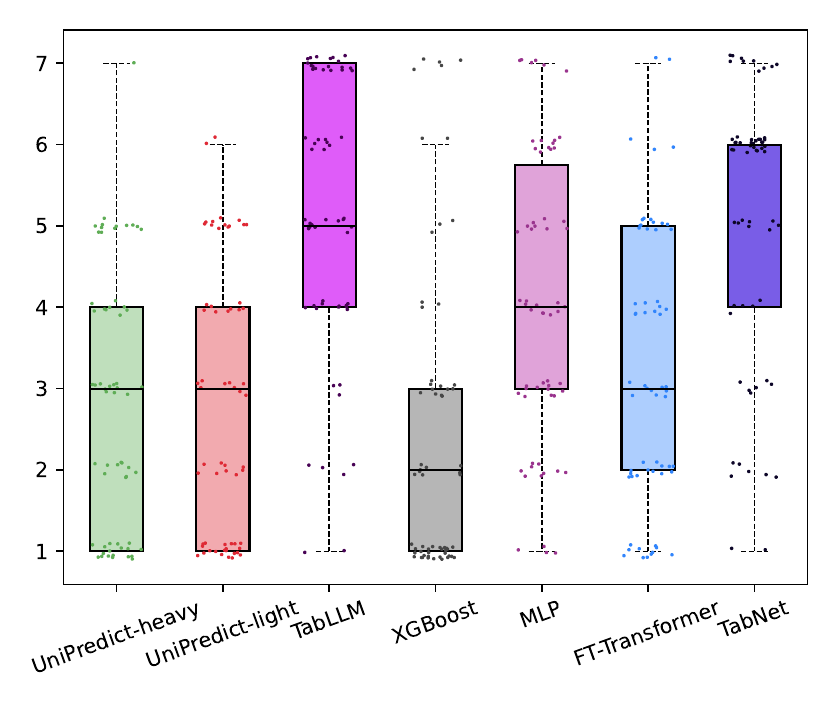}
         \caption{Model Rank (The lower the better)}
         \label{fig:few-shot-result-mid-rank}
     \end{subfigure}
     \caption{The average accuracy and rank of \texttt{UniPredict-heavy}, \texttt{UniPredict-light}, \texttt{TabLLM} \texttt{XGBoost}, \texttt{MLP}, \texttt{TabNet} and \texttt{FT-Transformer} on the few-shot dataset with train-set-ratio set to 0.5.}
    \label{fig:few-shot-result-mid}
\end{figure}

We present additional accuracy/ranking figures and datapoints for the few-shot datasets. Figure \ref{fig:few-shot-result-low} demonstrates each model's performance when train-set-proportion=0.1, Figure \ref{fig:few-shot-result-mid} shows their performance when the value is set to 0.5, and Figure \ref{fig:few-shot-result-low} gives the picture of models at a resource-rich setup (train-set-proportion=0.9). See Section \ref{sec:few-shot-results} for detailed discussion.

\fontsize{8}{11}\selectfont
\begin{longtabu} to \textwidth {p{2in}|ccccccccc}
\caption{The performance of \texttt{UniPredict-heavy} (\textbf{UniP-h}), its ablation (\textbf{Abl-h}), \texttt{UniPredict-light} (\textbf{UniP-l}), its ablation (\textbf{Abl-l}), \texttt{TabLLM} (\textbf{TabLLM}), \texttt{XGBoost} (\textbf{XGBoost}), \texttt{MLP} (\textbf{MLP}), \texttt{FT-Transformer} (\textbf{FT-Trans}), and \texttt{TabNet} (\textbf{TabNet}) on the supervised datasets. Each model's accuracy on the test set is reported. See Section \ref{sec:training-results} for the result analysis.} \\
\textbf{Dataset Name} & \textbf{UniP-h} & \textbf{Abl-h} & \textbf{UniP-l} & \textbf{Abl-l} & \textbf{TabLLM} & \textbf{XGBoost} & \textbf{MLP} & \textbf{FT-Trans} & \textbf{TabNet}\\
\hline
arnavsmayan-netflix-userbase-dataset & 0.632 & 0.556 & 0.616 & 0.596 & 0.332 & 0.600 & 0.372 & 0.608 & 0.564 \\
deependraverma13-diabetes-healthcare-comprehensive-dataset & 0.701 & 0.649 & 0.740 & 0.688 & 0.597 & 0.727 & 0.779 & 0.701 & 0.597 \\
bhanupratapbiswas-uber-data-analysis & 0.940 & 0.948 & 0.940 & 0.845 & 0.914 & 0.009 & 0.914 & 0.897 & 0.922 \\
swathiunnikrishnan-amazon-consumer-behaviour-dataset & 0.262 & 0.279 & 0.295 & 0.246 & 0.000 & 0.328 & 0.279 & 0.410 & 0.213 \\
hemanthhari-psycological-effects-of-covid & 0.763 & 0.737 & 0.822 & 0.356 & 0.000 & 0.805 & 0.000 & 0.153 & 0.000 \\
arslanr369-bitcoin-price-2014-2023 & 0.994 & 0.975 & 0.988 & 0.418 & 0.947 & 1.000 & 0.235 & 0.994 & 0.551 \\
saloni1712-chatgpt-app-reviews & 0.948 & 0.517 & 0.957 & 0.483 & 0.000 & 0.400 & 0.322 & 0.483 & 0.483 \\
naveenkumar20bps1137-predict-students-dropout-and-academic-success & 0.616 & 0.510 & 0.616 & 0.415 & 0.000 & 0.777 & 0.628 & 0.738 & 0.628 \\
sanjanchaudhari-user-behavior-on-instagram & 0.865 & 0.832 & 0.865 & 0.830 & 0.805 & 0.808 & 0.652 & 0.833 & 0.824 \\
bhanupratapbiswas-bollywood-actress-name-and-movie-list & 0.527 & 0.527 & 0.527 & 0.403 & 0.357 & 0.581 & 0.000 & 0.651 & 0.000 \\
arnavsmayan-vehicle-manufacturing-dataset & 0.350 & 0.305 & 0.395 & 0.325 & 0.240 & 0.000 & 0.000 & 0.000 & 0.000 \\
bharath011-heart-disease-classification-dataset & 0.970 & 0.652 & 0.962 & 0.568 & 0.561 & 0.000 & 0.000 & 0.000 & 0.000 \\
shroukgomaa-babies-food-ingredients & 0.600 & 0.886 & 0.671 & 0.400 & 0.243 & 0.986 & 0.000 & 0.286 & 0.000 \\
amirhosseinmirzaie-countries-life-expectancy & 0.804 & 0.782 & 0.804 & 0.435 & 0.505 & 0.905 & 0.000 & 0.277 & 0.000 \\
amirhosseinmirzaie-pistachio-types-detection & 0.855 & 0.808 & 0.872 & 0.762 & 0.669 & 0.895 & 0.407 & 0.866 & 0.407 \\
shashankshukla123123-marketing-campaign & 0.844 & 0.786 & 0.848 & 0.674 & 0.781 & 0.848 & 0.000 & 0.821 & 0.000 \\
uciml-pima-indians-diabetes-database & 0.727 & 0.675 & 0.701 & 0.610 & 0.597 & 0.727 & 0.779 & 0.688 & 0.597 \\
shubhamgupta012-titanic-dataset & 0.854 & 0.730 & 0.820 & 0.573 & 0.607 & 0.775 & 0.809 & 0.764 & 0.629 \\
bhanupratapbiswas-fashion-products & 0.380 & 0.340 & 0.320 & 0.350 & 0.280 & 0.390 & 0.350 & 0.440 & 0.310 \\
blastchar-telco-customer-churn & 0.834 & 0.749 & 0.827 & 0.732 & 0.743 & 0.728 & 0.447 & 0.762 & 0.789 \\
mirichoi0218-insurance & 0.851 & 0.843 & 0.866 & 0.575 & 0.440 & 0.821 & 0.746 & 0.881 & 0.455 \\
suraj520-dairy-goods-sales-dataset & 0.734 & 0.730 & 0.661 & 0.432 & 0.256 & 0.965 & 0.813 & 0.933 & 0.938 \\
uciml-red-wine-quality-cortez-et-al-2009 & 0.544 & 0.519 & 0.562 & 0.394 & 0.394 & 0.662 & 0.588 & 0.644 & 0.438 \\
akshaydattatraykhare-diabetes-dataset & 0.675 & 0.662 & 0.766 & 0.636 & 0.597 & 0.727 & 0.779 & 0.701 & 0.597 \\
arnabchaki-data-science-salaries-2023 & 0.963 & 0.971 & 0.963 & 0.763 & 0.902 & 0.995 & 0.588 & 0.963 & 0.258 \\
prkhrawsthi-bitcoin-usd-daily-price-with-volume-2015-2023 & 0.990 & 0.984 & 0.984 & 0.566 & 0.971 & 0.997 & 0.238 & 0.981 & 0.559 \\
hawkingcr-airbnb-for-boston-with-fraud-detection & 0.886 & 0.836 & 0.889 & 0.752 & 0.000 & 0.864 & 0.850 & 0.833 & 0.866 \\
saunakghosh-nba-players-dataset & 0.856 & 0.850 & 0.850 & 0.811 & 0.840 & 0.875 & 0.000 & 0.115 & 0.000 \\
rtatman-chocolate-bar-ratings & 0.400 & 0.272 & 0.372 & 0.278 & 0.306 & 0.350 & 0.333 & 0.344 & 0.311 \\
pavansubhasht-ibm-hr-analytics-attrition-dataset & 0.871 & 0.810 & 0.830 & 0.755 & 0.000 & 0.857 & 0.769 & 0.776 & 0.837 \\
gyanprakashkushwaha-laptop-price-prediction-cleaned-dataset & 0.633 & 0.539 & 0.609 & 0.477 & 0.211 & 0.758 & 0.586 & 0.656 & 0.438 \\
fedesoriano-stroke-prediction-dataset & 1.000 & 0.937 & 1.000 & 0.916 & 0.914 & 0.937 & 0.000 & 0.945 & 0.000 \\
bhanupratapbiswas-world-top-billionaires & 0.989 & 0.859 & 0.962 & 0.466 & 0.435 & 0.996 & 0.557 & 0.969 & 0.000 \\
vstacknocopyright-blood-transfusion-service-center-data & 0.960 & 0.693 & 0.973 & 0.653 & 0.667 & 0.747 & 0.800 & 0.773 & 0.187 \\
ashishkumarjayswal-movies-updated-data & 0.295 & 0.292 & 0.233 & 0.212 & 0.000 & 0.410 & 0.000 & 0.003 & 0.000 \\
bhanupratapbiswas-ipl-dataset-2008-2016 & 0.121 & 0.345 & 0.207 & 0.121 & 0.000 & 0.966 & 0.000 & 0.000 & 0.000 \\
mathchi-diabetes-data-set & 0.701 & 0.662 & 0.753 & 0.688 & 0.597 & 0.727 & 0.779 & 0.688 & 0.597 \\
harishkumardatalab-medical-insurance-price-prediction & 0.795 & 0.809 & 0.773 & 0.471 & 0.647 & 0.986 & 0.788 & 0.899 & 0.719 \\
arslanr369-roblox-stock-pricing-2021-2023 & 0.966 & 0.966 & 0.966 & 0.776 & 0.241 & 1.000 & 0.345 & 1.000 & 0.241 \\
yasserh-titanic-dataset & 0.789 & 0.800 & 0.722 & 0.767 & 0.533 & 0.511 & 0.000 & 0.600 & 0.000 \\
iqmansingh-company-employee-dataset & 0.886 & 0.698 & 0.908 & 0.524 & 0.384 & 0.922 & 0.230 & 0.928 & 0.838 \\
shivamb-disney-movies-and-tv-shows & 1.000 & 1.000 & 1.000 & 1.000 & 1.000 & 0.986 & 0.690 & 0.966 & 0.434 \\
alexisbcook-pakistan-intellectual-capital & 0.843 & 0.930 & 0.696 & 0.183 & 0.000 & 0.974 & 0.000 & 0.035 & 0.000 \\
tahzeer-indian-startups-by-state & 0.670 & 0.618 & 0.707 & 0.493 & 0.479 & 0.624 & 0.285 & 0.534 & 0.292 \\
harshitshankhdhar-imdb-dataset-of-top-1000-movies-and-tv-shows & 0.250 & 0.320 & 0.330 & 0.240 & 0.260 & 0.430 & 0.000 & 0.290 & 0.000 \\
shreyapurohit-anime-data & 0.969 & 0.991 & 0.969 & 0.626 & 0.988 & 0.270 & 0.696 & 0.990 & 0.987 \\
raddar-icr-integer-data & 0.000 & 0.000 & 0.694 & 0.484 & 0.000 & 0.968 & 0.000 & 0.790 & 0.000 \\
uciml-mushroom-classification & 1.000 & 1.000 & 0.999 & 0.999 & 0.998 & 1.000 & 1.000 & 1.000 & 1.000 \\
adityakadiwal-water-potability & 0.631 & 0.503 & 0.567 & 0.421 & 0.500 & 0.692 & 0.000 & 0.622 & 0.000 \\
shreyanshverma27-imdb-horror-chilling-movie-dataset & 0.286 & 0.321 & 0.274 & 0.274 & 0.190 & 0.262 & 0.298 & 0.357 & 0.286 \\
ruchi798-data-science-job-salaries & 0.836 & 0.803 & 0.754 & 0.607 & 0.344 & 0.951 & 0.246 & 0.967 & 0.295 \\
hesh97-titanicdataset-traincsv & 0.778 & 0.733 & 0.756 & 0.733 & 0.533 & 0.511 & 0.000 & 0.600 & 0.000 \\
phangud-spamcsv & 1.000 & 0.995 & 1.000 & 0.989 & 0.993 & 0.864 & 0.869 & 0.869 & 0.869 \\
dileep070-heart-disease-prediction-using-logistic-regression & 0.965 & 0.767 & 0.962 & 0.762 & 0.743 & 0.823 & 0.000 & 0.851 & 0.000 \\
abcsds-pokemon & 0.087 & 0.113 & 0.100 & 0.037 & 0.000 & 0.300 & 0.225 & 0.312 & 0.000 \\
atharvaingle-crop-recommendation-dataset & 0.973 & 0.964 & 0.914 & 0.600 & 0.000 & 0.995 & 0.973 & 0.991 & 0.132 \\
rounakbanik-pokemon & 1.000 & 0.975 & 0.975 & 0.852 & 0.000 & 1.000 & 0.000 & 0.864 & 0.000 \\
thedevastator-cancer-patients-and-air-pollution-a-new-link & 0.450 & 0.390 & 0.560 & 0.310 & 0.220 & 0.560 & 0.360 & 0.670 & 0.290 \\
andrewmvd-fetal-health-classification & 0.859 & 0.798 & 0.845 & 0.620 & 0.000 & 0.958 & 0.845 & 0.925 & 0.479 \\
saurabh00007-diabetescsv & 0.753 & 0.662 & 0.714 & 0.662 & 0.597 & 0.727 & 0.779 & 0.649 & 0.597 \\
larsen0966-student-performance-data-set & 0.138 & 0.354 & 0.123 & 0.062 & 0.000 & 0.446 & 0.262 & 0.462 & 0.015 \\
nikhil1e9-netflix-stock-price & 0.989 & 0.994 & 0.991 & 0.593 & 0.994 & 1.000 & 0.358 & 0.981 & 0.829 \\
yasserh-wine-quality-dataset & 0.504 & 0.504 & 0.557 & 0.400 & 0.357 & 0.617 & 0.530 & 0.583 & 0.348 \\
ashishkumarjayswal-loanamount-approval & 0.758 & 0.694 & 0.758 & 0.677 & 0.677 & 0.677 & 0.000 & 0.339 & 0.000 \\
ananthr1-weather-prediction & 0.925 & 0.707 & 0.952 & 0.694 & 0.333 & 0.837 & 0.714 & 0.837 & 0.476 \\
thedevastator-higher-education-predictors-of-student-retention & 0.862 & 0.856 & 0.885 & 0.833 & 0.000 & 0.894 & 0.885 & 0.874 & 0.907 \\
rpaguirre-tesla-stock-price & 0.976 & 0.976 & 0.982 & 0.888 & 0.965 & 1.000 & 0.235 & 0.971 & 0.453 \\
muhammadtsabitulazmi-liga-1-indonesia-player-dataset & 0.105 & 0.105 & 0.105 & 0.105 & 0.088 & 0.228 & 0.175 & 0.193 & 0.070 \\
ashishkumarjayswal-diabetes-dataset & 0.701 & 0.675 & 0.727 & 0.623 & 0.597 & 0.727 & 0.779 & 0.675 & 0.597 \\
wearefuture01-hepatitis-c-prediction & 0.935 & 0.790 & 0.887 & 0.516 & 0.855 & 0.984 & 0.000 & 0.113 & 0.000 \\
aakashjoshi123-exercise-and-fitness-metrics-dataset & 0.801 & 0.809 & 0.796 & 0.607 & 0.253 & 0.804 & 0.442 & 0.796 & 0.770 \\
kumargh-pimaindiansdiabetescsv & 0.104 & 0.143 & 0.169 & 0.130 & 0.065 & 0.156 & 0.182 & 0.169 & 0.078 \\
gauravduttakiit-resume-dataset & 0.144 & 0.186 & 0.124 & 0.144 & 0.000 & 0.021 & 0.124 & 0.340 & 0.031 \\
surajjha101-stores-area-and-sales-data & 0.289 & 0.300 & 0.233 & 0.222 & 0.300 & 0.233 & 0.178 & 0.267 & 0.222 \\
rishikeshkonapure-hr-analytics-prediction & 0.850 & 0.789 & 0.857 & 0.741 & 0.762 & 0.850 & 0.769 & 0.898 & 0.837 \\
eishkaran-heart-disease & 0.866 & 0.815 & 0.874 & 0.723 & 0.681 & 0.966 & 0.866 & 0.933 & 0.655 \\
vikramamin-customer-churn-decision-tree-and-random-forest & 0.834 & 0.749 & 0.837 & 0.694 & 0.743 & 0.799 & 0.447 & 0.755 & 0.789 \\
redwankarimsony-heart-disease-data & 0.489 & 0.446 & 0.543 & 0.435 & 0.359 & 0.587 & 0.000 & 0.380 & 0.000 \\
hashemi221022-diabetes & 0.675 & 0.636 & 0.753 & 0.701 & 0.597 & 0.727 & 0.779 & 0.675 & 0.597 \\
rajyellow46-wine-quality & 0.469 & 0.412 & 0.489 & 0.366 & 0.406 & 0.691 & 0.000 & 0.002 & 0.000 \\
vikramamin-time-series-forecasting-using-prophet-in-r & 0.317 & 0.738 & 0.317 & 0.552 & 0.306 & 0.344 & 0.257 & 0.273 & 0.262 \\
reihanenamdari-breast-cancer & 0.397 & 0.308 & 0.400 & 0.345 & 0.261 & 0.367 & 0.345 & 0.347 & 0.355 \\
uciml-indian-liver-patient-records & 0.695 & 0.610 & 0.763 & 0.627 & 0.678 & 0.712 & 0.000 & 0.763 & 0.000 \\
teertha-ushealthinsurancedataset & 0.903 & 0.836 & 0.851 & 0.657 & 0.440 & 0.821 & 0.746 & 0.843 & 0.455 \\
ninzaami-loan-predication & 0.694 & 0.677 & 0.790 & 0.581 & 0.677 & 0.710 & 0.000 & 0.339 & 0.000 \\
timoboz-tesla-stock-data-from-2010-to-2020 & 0.983 & 0.992 & 0.979 & 0.409 & 0.909 & 0.240 & 0.273 & 0.963 & 0.517 \\
elakiricoder-gender-classification-dataset & 0.976 & 0.972 & 0.972 & 0.964 & 0.946 & 0.970 & 0.964 & 0.976 & 0.978 \\
jainilcoder-netflix-stock-price-prediction & 0.950 & 0.960 & 0.970 & 0.960 & 0.267 & 1.000 & 0.257 & 0.970 & 0.257 \\
burak3ergun-loan-data-set & 0.710 & 0.661 & 0.742 & 0.742 & 0.677 & 0.677 & 0.000 & 0.339 & 0.000 \\
sanjanchaudhari-bankloan & 0.600 & 0.613 & 0.667 & 0.580 & 0.527 & 0.700 & 0.573 & 0.687 & 0.727 \\
alirezachahardoli-bank-personal-loan-1 & 0.982 & 0.982 & 0.992 & 0.974 & 0.954 & 0.984 & 0.892 & 0.980 & 0.982 \\
sbhatti-financial-sentiment-analysis & 1.000 & 0.750 & 1.000 & 0.749 & 0.737 & 0.491 & 0.321 & 0.533 & 0.533 \\
altruistdelhite04-gold-price-data & 0.917 & 0.921 & 0.900 & 0.860 & 0.258 & 0.939 & 0.703 & 0.956 & 0.489 \\
carolzhangdc-imdb-5000-movie-dataset & 0.453 & 0.370 & 0.424 & 0.311 & 0.000 & 0.552 & 0.000 & 0.255 & 0.000 \\
desalegngeb-german-fintech-companies & 0.969 & 0.929 & 0.929 & 0.867 & 0.000 & 1.000 & 0.000 & 0.143 & 0.000 \\
crxxom-manhwa-dataset & 0.993 & 0.990 & 0.990 & 0.936 & 0.000 & 0.990 & 0.000 & 0.366 & 0.000 \\
varpit94-tesla-stock-data-updated-till-28jun2021 & 0.990 & 0.986 & 0.990 & 0.328 & 0.943 & 0.530 & 0.216 & 0.983 & 0.693 \\
hashemi221022-bank-loans & 0.986 & 0.990 & 0.986 & 0.970 & 0.954 & 0.984 & 0.892 & 0.984 & 0.982 \\
geomack-spotifyclassification & 1.000 & 0.995 & 0.995 & 0.554 & 0.941 & 1.000 & 0.921 & 0.990 & 0.861 \\
jillanisofttech-brain-stroke-dataset & 1.000 & 0.920 & 1.000 & 0.910 & 0.912 & 0.928 & 0.938 & 0.906 & 0.938 \\
mayankpatel14-second-hand-used-cars-data-set-linear-regression & 0.740 & 0.730 & 0.760 & 0.520 & 0.320 & 0.190 & 0.790 & 0.910 & 0.270 \\
rkiattisak-student-performance-in-mathematics & 0.530 & 0.420 & 0.550 & 0.370 & 0.290 & 0.620 & 0.580 & 0.660 & 0.350 \\
sabasaeed1953-stock-prices-of-2023 & 0.957 & 0.986 & 0.957 & 0.571 & 0.357 & 0.957 & 0.257 & 0.957 & 0.229 \\
primaryobjects-voicegender & 0.953 & 0.962 & 0.965 & 0.536 & 0.934 & 0.019 & 0.972 & 0.987 & 0.669 \\
maryammanoochehry-bank-personal-loan & 0.986 & 0.988 & 0.988 & 0.972 & 0.954 & 0.984 & 0.892 & 0.982 & 0.982 \\
bhavkaur-simplified-titanic-dataset & 0.982 & 0.737 & 0.978 & 0.701 & 0.723 & 0.768 & 0.719 & 0.754 & 0.750 \\
sidhus-crab-age-prediction & 0.621 & 0.500 & 0.669 & 0.431 & 0.415 & 0.585 & 0.595 & 0.610 & 0.608 \\
ahsan81-superstore-marketing-campaign-dataset & 0.893 & 0.835 & 0.835 & 0.750 & 0.768 & 0.884 & 0.000 & 0.862 & 0.000 \\
fedesoriano-hepatitis-c-dataset & 0.952 & 0.790 & 0.871 & 0.435 & 0.855 & 0.984 & 0.000 & 0.113 & 0.000 \\
oles04-bundesliga-seasons & 1.000 & 1.000 & 1.000 & 1.000 & 0.000 & 1.000 & 0.000 & 0.584 & 0.000 \\
gabrielsantello-cars-purchase-decision-dataset & 0.930 & 0.810 & 0.830 & 0.690 & 0.460 & 0.900 & 0.440 & 0.910 & 0.400 \\
andrewmvd-udemy-courses & 0.908 & 0.984 & 0.913 & 0.927 & 0.970 & 0.207 & 0.402 & 0.454 & 0.000 \\
whenamancodes-students-performance-in-exams & 0.620 & 0.500 & 0.640 & 0.350 & 0.260 & 0.600 & 0.630 & 0.580 & 0.310 \\
patelprashant-employee-attrition & 0.844 & 0.830 & 0.850 & 0.714 & 0.762 & 0.844 & 0.769 & 0.878 & 0.837 \\
barun2104-telecom-churn & 0.892 & 0.862 & 0.910 & 0.793 & 0.805 & 0.913 & 0.853 & 0.898 & 0.865 \\
kandij-diabetes-dataset & 0.727 & 0.675 & 0.740 & 0.688 & 0.597 & 0.727 & 0.779 & 0.727 & 0.597 \\
vedavyasv-usa-housing & 0.670 & 0.634 & 0.670 & 0.564 & 0.272 & 0.226 & 0.222 & 0.706 & 0.700 \\
team-ai-spam-text-message-classification & 1.000 & 0.995 & 1.000 & 0.989 & 0.993 & 0.864 & 0.869 & 0.869 & 0.869 \\
prevek18-ames-housing-dataset & 0.683 & 0.614 & 0.710 & 0.491 & 0.000 & 0.823 & 0.000 & 0.256 & 0.000 \\
mazlumi-ielts-writing-scored-essays-dataset & 0.542 & 0.333 & 0.597 & 0.312 & 0.000 & 0.472 & 0.000 & 0.243 & 0.000 \\
vijayvvenkitesh-microsoft-stock-time-series-analysis & 0.993 & 0.980 & 0.993 & 0.500 & 0.375 & 0.987 & 0.283 & 0.993 & 0.217 \\
ruchi798-tv-shows-on-netflix-prime-video-hulu-and-disney & 0.534 & 0.451 & 0.549 & 0.395 & 0.426 & 0.084 & 0.432 & 0.480 & 0.436 \\
tarkkaanko-amazon & 0.974 & 0.746 & 0.982 & 0.711 & 0.000 & 0.793 & 0.768 & 0.715 & 0.789 \\
kingabzpro-cosmetics-datasets & 0.318 & 0.757 & 0.284 & 0.581 & 0.000 & 0.372 & 0.162 & 0.311 & 0.203 \\
receplyasolu-6k-weather-labeled-spotify-songs & 0.319 & 0.339 & 0.308 & 0.301 & 0.218 & 0.443 & 0.185 & 0.327 & 0.316 \\
kabure-german-credit-data-with-risk & 0.750 & 0.640 & 0.700 & 0.680 & 0.600 & 0.790 & 0.710 & 0.680 & 0.490 \\
mahnazarjmand-bank-personal-loan & 0.992 & 0.990 & 0.982 & 0.968 & 0.946 & 0.992 & 0.904 & 0.992 & 0.984 \\
sudarshan6561-ipl-2023 & 0.368 & 0.281 & 0.474 & 0.211 & 0.333 & 0.667 & 0.000 & 0.211 & 0.000 \\
agirlcoding-all-space-missions-from-1957 & 0.988 & 0.887 & 0.977 & 0.873 & 0.864 & 0.889 & 0.813 & 0.855 & 0.898 \\
mfaisalqureshi-spam-email & 1.000 & 0.991 & 1.000 & 0.989 & 0.993 & 0.864 & 0.869 & 0.869 & 0.869 \\
cpluzshrijayan-milkquality & 0.991 & 0.906 & 0.925 & 0.896 & 0.481 & 1.000 & 0.849 & 1.000 & 0.274 \\
awaiskaggler-insurance-csv & 0.649 & 0.687 & 0.664 & 0.575 & 0.216 & 0.888 & 0.239 & 0.821 & 0.209 \\
thedevastator-employee-attrition-and-factors & 0.830 & 0.803 & 0.844 & 0.762 & 0.762 & 0.850 & 0.769 & 0.884 & 0.837 \\
surajjha101-top-youtube-channels-data & 0.140 & 0.170 & 0.210 & 0.140 & 0.130 & 0.300 & 0.230 & 0.200 & 0.140 \\
hansrobertson-american-companies-profits-and-benefits-from-ai & 0.331 & 0.303 & 0.331 & 0.324 & 0.269 & 0.276 & 0.255 & 0.310 & 0.283 \\
dansbecker-aer-credit-card-data & 0.977 & 0.962 & 0.985 & 0.947 & 0.970 & 0.970 & 0.977 & 0.939 & 0.712 \\
whenamancodes-predict-diabities & 0.727 & 0.662 & 0.740 & 0.636 & 0.597 & 0.727 & 0.779 & 0.714 & 0.597 \\
nancyalaswad90-review & 0.701 & 0.675 & 0.727 & 0.597 & 0.597 & 0.727 & 0.779 & 0.727 & 0.597 \\
ruchi798-student-feedback-survey-responses & 0.089 & 0.059 & 0.079 & 0.089 & 0.099 & 0.069 & 0.079 & 0.109 & 0.129 \\
siddharthss-crop-recommendation-dataset & 0.968 & 0.955 & 0.932 & 0.564 & 0.041 & 0.995 & 0.973 & 0.986 & 0.132 \\
therealsampat-predict-movie-success-rate & 0.905 & 0.833 & 0.929 & 0.595 & 0.714 & 1.000 & 0.000 & 0.798 & 0.000 \\
maryalebron-life-expectancy-data & 0.235 & 0.279 & 0.269 & 0.231 & 0.313 & 0.241 & 0.000 & 0.286 & 0.000 \\
noordeen-insurance-premium-prediction & 0.910 & 0.858 & 0.873 & 0.590 & 0.343 & 0.836 & 0.754 & 0.843 & 0.485 \\
ybifoundation-food-app-business & 0.000 & 0.000 & 0.199 & 0.090 & 0.140 & 0.430 & 0.213 & 0.416 & 0.199 \\
oles04-top-leagues-player & 0.385 & 0.336 & 0.412 & 0.286 & 0.286 & 0.233 & 0.000 & 0.160 & 0.000 \\
buntyshah-auto-insurance-claims-data & 0.810 & 0.780 & 0.750 & 0.770 & 0.710 & 0.790 & 0.000 & 0.770 & 0.000 \\
lightonkalumba-us-womens-labor-force-participation & 1.000 & 0.987 & 1.000 & 0.947 & 0.789 & 1.000 & 1.000 & 1.000 & 0.592 \\
tejashvi14-employee-future-prediction & 0.929 & 0.790 & 0.940 & 0.732 & 0.682 & 0.865 & 0.633 & 0.848 & 0.345 \\
arnabchaki-indian-restaurants-2023 & 0.406 & 0.314 & 0.403 & 0.320 & 0.332 & 0.412 & 0.276 & 0.341 & 0.376 \\
kanths028-usa-housing & 0.624 & 0.646 & 0.692 & 0.566 & 0.272 & 0.226 & 0.222 & 0.692 & 0.700 \\
ravibarnawal-mutual-funds-india-detailed & 0.280 & 0.305 & 0.268 & 0.244 & 0.195 & 0.427 & 0.000 & 0.146 & 0.000 \\
dsfelix-us-stores-sales & 0.974 & 0.960 & 0.967 & 0.791 & 0.826 & 0.998 & 0.809 & 0.979 & 0.880 \\
sanjanchaudhari-netflix-dataset & 0.319 & 0.467 & 0.423 & 0.280 & 0.154 & 0.538 & 0.231 & 0.385 & 0.198 \\
tejashvi14-engineering-placements-prediction & 0.956 & 0.771 & 0.946 & 0.764 & 0.781 & 0.879 & 0.822 & 0.869 & 0.586 \\
bhavkaur-hotel-guests-dataset & 0.970 & 0.800 & 0.970 & 0.740 & 0.770 & 0.845 & 0.000 & 0.855 & 0.000 \\
warcoder-earthquake-dataset & 0.835 & 0.835 & 0.962 & 0.241 & 0.266 & 0.759 & 0.418 & 0.722 & 0.253 \\
mayurdalvi-simple-linear-regression-placement-data & 0.650 & 0.520 & 0.690 & 0.510 & 0.530 & 0.550 & 0.470 & 0.610 & 0.500 \\
arashnic-time-series-forecasting-with-yahoo-stock-price & 0.995 & 0.967 & 0.989 & 0.530 & 0.760 & 0.262 & 0.262 & 1.000 & 0.273 \\
bretmathyer-telemedicine-used & 0.934 & 0.931 & 0.934 & 0.737 & 0.830 & 0.988 & 0.000 & 0.481 & 0.000 \\
iamsumat-spotify-top-2000s-mega-dataset & 0.375 & 0.305 & 0.340 & 0.340 & 0.295 & 0.345 & 0.310 & 0.355 & 0.245 \\
ahsan81-food-ordering-and-delivery-app-dataset & 0.521 & 0.289 & 0.547 & 0.253 & 0.295 & 0.353 & 0.147 & 0.379 & 0.389 \\
kreeshrajani-human-stress-prediction & 0.690 & 0.764 & 0.715 & 0.736 & 0.687 & 0.602 & 0.549 & 0.556 & 0.549 \\
shivamb-hm-stores-dataset & 0.644 & 0.530 & 0.633 & 0.481 & 0.481 & 0.605 & 0.000 & 0.037 & 0.000 \\
christinestevens-cstevens-peloton-data & 0.179 & 0.209 & 0.238 & 0.168 & 0.171 & 0.559 & 0.000 & 0.150 & 0.000 \\
aakashjoshi123-spotify-top-hits-data & 0.690 & 0.690 & 0.690 & 0.670 & 0.620 & 0.780 & 0.000 & 0.740 & 0.000 \\
ishadss-productivity-prediction-of-garment-employees & 0.475 & 0.442 & 0.558 & 0.358 & 0.250 & 0.683 & 0.000 & 0.242 & 0.000 \\
chirin-africa-economic-banking-and-systemic-crisis-data & 0.972 & 0.981 & 0.991 & 0.887 & 0.877 & 0.991 & 0.934 & 0.991 & 0.896 \\
mayuriawati-bangalore-chain-restaurants-ratings-and-reviews & 0.814 & 0.940 & 0.776 & 0.617 & 0.093 & 1.000 & 0.131 & 0.934 & 0.186 \\
azminetoushikwasi--lionel-messi-all-club-goals & 0.704 & 0.662 & 0.563 & 0.493 & 0.423 & 0.662 & 0.634 & 0.606 & 0.056 \\
\label{tab:supervised-results}
\end{longtabu}

\fontsize{10}{8}\selectfont

\subsection{Example Cause of Failure}\label{apdx: failure}

\begin{figure}
     \centering
     \begin{subfigure}[b]{0.45\textwidth}
         \centering
         \includegraphics[width=\textwidth]{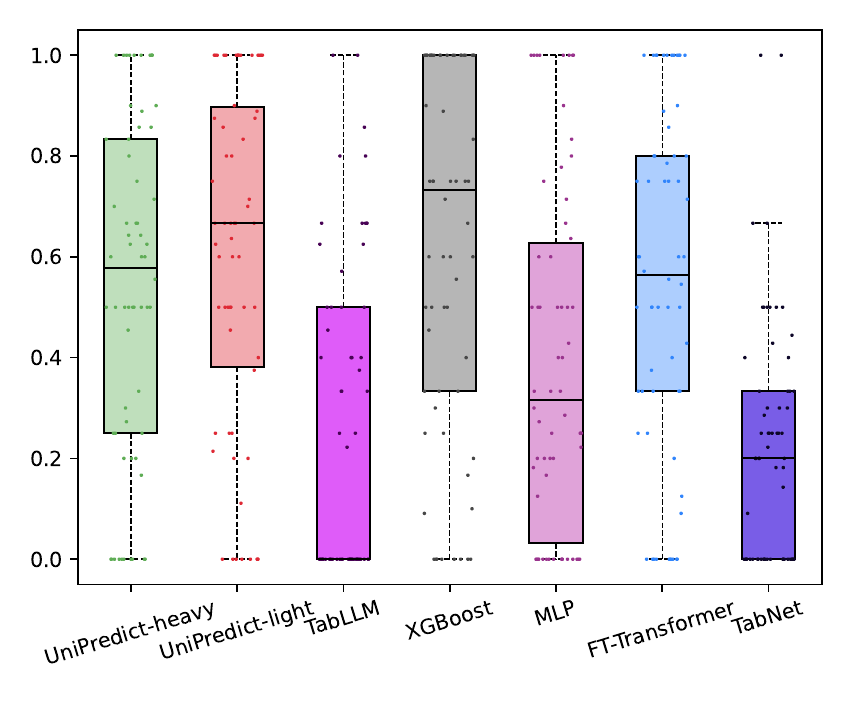}
         \caption{Model Accuracy (The higher the better)}
         \label{fig:few-shot-result-high-accuracy}
     \end{subfigure}
     \hfill
     \begin{subfigure}[b]{0.45\textwidth}
         \centering
         \includegraphics[width=\textwidth]{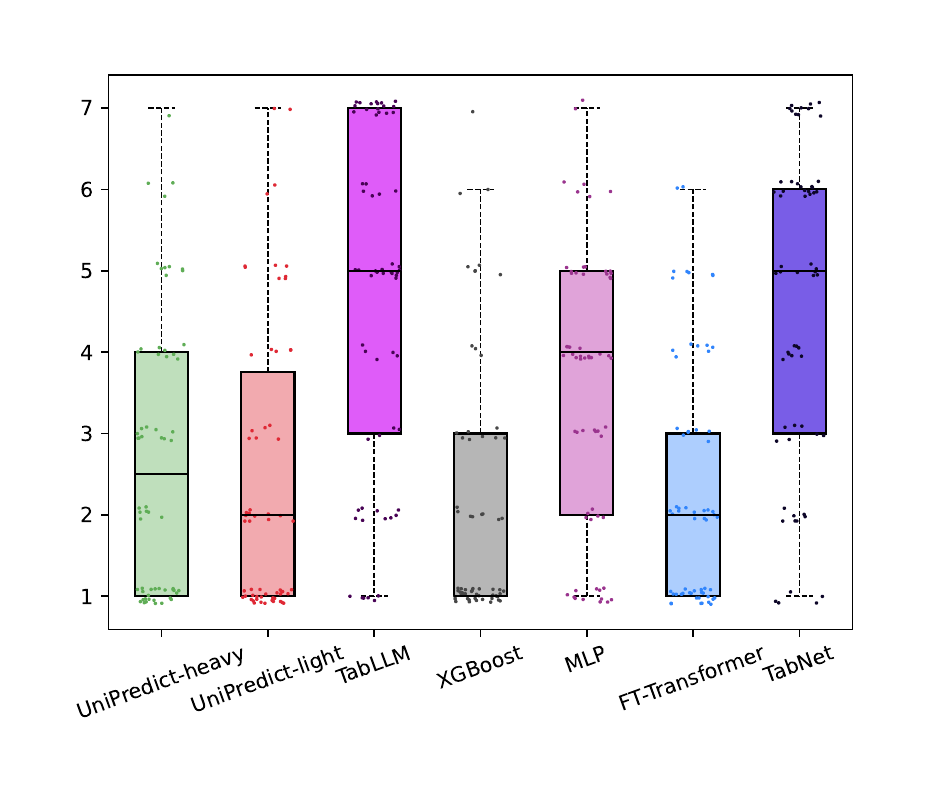}
         \caption{Model Rank (The lower the better)}
         \label{fig:few-shot-result-high-rank}
     \end{subfigure}
     \caption{The average accuracy and rank of \texttt{UniPredict-heavy}, \texttt{UniPredict-light}, \texttt{TabLLM} \texttt{XGBoost}, \texttt{MLP}, \texttt{TabNet} and \texttt{FT-Transformer} on the few-shot dataset with train-set-ratio set to 0.9.}
    \label{fig:few-shot-result-high}
\end{figure}

In Section \ref{sec:case-study} we presented a failure study on \method, and gave some possible issues that cause the model to give poor performance. We present demonstrations for each causes below:
\begin{lstlisting}[caption={Example columns and values that have the \textbf{COL} (too many column) problem. Data origin: \texttt{suraj520-dairy-goods-sales-dataset}}, language=python]

# Column names: 
"""
Marital status,Application mode,Application order,Course,Daytime/evening attendance,Previous qualification,Nacionality,Mother's qualification,Father's qualification,Mother's occupation,Father's occupation,Displaced,Educational special needs,Debtor,Tuition fees up to date,Gender,Scholarship holder,Age at enrollment,International,Curricular units 1st sem (credited),Curricular units 1st sem (enrolled),Curricular units 1st sem (evaluations),Curricular units 1st sem (approved),Curricular units 1st sem (grade),Curricular units 1st sem (without evaluations),Curricular units 2nd sem (credited),Curricular units 2nd sem (enrolled),Curricular units 2nd sem (evaluations),Curricular units 2nd sem (approved),Curricular units 2nd sem (grade),Curricular units 2nd sem (without evaluations),Unemployment rate,Inflation rate,GDP,Target
"""

# Column value example:
"""
1,8,5,2,1,1,1,13,10,6,10,1,0,0,1,1,0,20,0,0,0,0,0,0.0,0,0,0,0,0,0.0,0, 10.8,1.4,1.74,Dropout
"""

# Note: This sample also has the FV (Poorly represented Feature Values) problem as there are too many numerical values inside.
\end{lstlisting}

\begin{lstlisting}[caption={Example columns and values that have the \textbf{FV} (Poorly represented Feature Values) problem. Dataset origin: \texttt{yasserh-wine-quality-dataset}}, language=python]

# Column names: 
"""
fixed acidity,volatile acidity,citric acid,residual sugar,chlorides,free sulfur dioxide,total sulfur dioxide,density,pH,sulphates,alcohol,quality,Id
"""

# Column value example:
"""
7.4,0.7,0.0,1.9,0.076,11.0,34.0,0.9978,3.51,0.56,9.4,5,0
"""
\end{lstlisting}

\begin{lstlisting}[caption={Example columns and values that have the \textbf{META} (Inadequate or ambiguous Metadata) problem. Dataset origin: \texttt{kumargh-pimaindiansdiabetescsv}}, language=python]
# Dataset metadata:
"""
(No metadata)
"""

# Column names: 
"""
(No Column names)
"""

# Column value example:
"""
1,85,66,29,0,26.6,0.351,31,0
"""
\end{lstlisting}

\begin{lstlisting}[caption={Example columns and values that have the \textbf{OTH} (Other factors) problem. Dataset origin: \texttt{rkiattisak-student-performance-in-mathematics}}, language=python]
# Dataset metadata:
"""
Description: This dataset contains information on the performance of high school students in mathematics, including their grades and demographic information. The data was collected from three high schools in the United States.\n\n
Columns:\n\n\t
**Gender:** The gender of the student (male/female)\n\n\t
**Race/ethnicity:** The student's racial or ethnic background (Asian, African-American, Hispanic, etc.)\n\n\t
**Parental level of education:** The highest level of education attained by the student's parent(s) or guardian(s)\n\n\t
**Lunch:** Whether the student receives free or reduced-price lunch (yes/no)\n\n\t
**Test preparation course:** Whether the student completed a test preparation course (yes/no)\n\n\t
**Math score:** The student's score on a standardized mathematics test\n\n\t
**Reading score:** The student's score on a standardized reading test\n\n\t
**Writing score:** The student's score on a standardized writing test\n\nThis dataset could be used for various research questions related to education, such as examining the impact of parental education or test preparation courses on student performance. It could also be used to develop machine learning models to predict student performance based on demographic and other factors.\n\n
source: http://roycekimmons.com/tools/generated_data/exams\n"
"""

# Column names: 
"""
"gender","race/ethnicity","parental level of education","lunch","test preparation course","math score","reading score","writing score"
"""

# Column value example:
"""
"female","group D","some college","standard","completed","59","70","78"
"""

# Nothing is explicitly wrong in this dataset.
\end{lstlisting}

\end{document}